\documentclass{article}

\usepackage{microtype}
\usepackage{graphicx}
\usepackage{subcaption}
\usepackage{booktabs} 

\usepackage{hyperref}
\usepackage{multirow}

\usepackage[preprint]{icml2026}

\usepackage{amsmath}
\usepackage{amssymb}
\usepackage{mathtools}
\usepackage{amsthm}

\usepackage[capitalize,noabbrev]{cleveref}

\theoremstyle{plain}

\theoremstyle{definition}

\theoremstyle{remark}

\usepackage[disable,textsize=tiny]{todonotes}
\usepackage{placeins} 
\usepackage{float}
\icmltitlerunning{Drive-KD: Multi-Teacher Distillation for VLMs in Autonomous Driving} 

\begin{document}

\twocolumn[
  \icmltitle{Drive-KD: Multi-Teacher Distillation for VLMs in Autonomous Driving}

  \icmlsetsymbol{equal}{*}
    \icmlsetsymbol{lead}{\ensuremath{\dagger}}
    \icmlsetsymbol{corrauth}{\ensuremath{\ddagger}}
\begin{icmlauthorlist}
  \icmlauthor{Weitong Lian}{equal,zju}
  \icmlauthor{Zecong Tang}{equal,lead,zju}
  \icmlauthor{Haoran Li}{equal,zju}
  \icmlauthor{Tianjian Gao}{zju}
  \icmlauthor{Yifei Wang}{zju}
  \\ \vspace{0.2ex}
  \icmlauthor{Zixu Wang}{zju}
  \icmlauthor{Lingyi Meng}{zju}
  \icmlauthor{Tengju Ru}{zju}
  \icmlauthor{Zhejun Cui}{zju}
  \icmlauthor{Yichen Zhu}{zju}
  \icmlauthor{Hangshuo Cao}{zju}
  \\ \vspace{0.2ex}
  \icmlauthor{Qi Kang}{zju}
  \icmlauthor{Tianxing Chen}{hku}
  \icmlauthor{Kaixuan Wang}{hku}
  \icmlauthor{Yu Zhang}{corrauth,zju}
\end{icmlauthorlist}
  \icmlaffiliation{zju}{Zhejiang University, Hangzhou, China}
  \icmlaffiliation{hku}{The University of Hong Kong, Hong Kong, China}

  \icmlkeywords{Machine Learning, ICML}

  \vskip 0.3in
]

\makeatletter
\newcommand{\icmlnoticeblock}[1]{%
  {%
    \renewcommand\@makefnmark{}
    \renewcommand\@makefntext[1]{\noindent\hspace*{0pt}##1}
    \setlength{\parindent}{0pt}%
    \setlength{\leftskip}{0pt}%
    \setlength{\parskip}{0pt}%
    #1%
  }%
}
\makeatother
\printAffiliationsAndNotice{%
  \icmlnoticeblock{%
    \textsuperscript{*}\,Equal contribution.\par
    \textsuperscript{\ensuremath{\dagger}}\,Project leader.\par
    \textsuperscript{\ensuremath{\ddagger}}\,Corresponding author.\par
  }%
}


\begin{abstract}
  Autonomous driving is a safety-critical task, and recent advances in LLMs/VLMs have opened new possibilities in this domain. However, large models demand substantial GPU memory and exhibit high inference latency, while conventional supervised fine-tuning (SFT) often struggles to bridge the capability gaps of small models. To address these limitations, we propose Drive-KD, a framework that decomposes autonomous driving into a ``perception--reasoning--planning'' triad and transfers these capabilities via knowledge distillation to improve VLM performance on comprehensive driving VQA tasks. Through a systematic study of distillation design for autonomous driving, we identify layer-specific attention as the distillation signal to construct capability-specific single-teacher models. Moreover, we unify these single-teacher settings into a multi-teacher distillation framework and introduce asymmetric gradient projection to mitigate cross-capability gradient conflicts. Extensive evaluations validate the generalization of our method across diverse model families and scales. Experiments show that our distilled InternVL3-1B model, with $\sim$42$\times$ less GPU memory and $\sim$11.4$\times$ higher throughput, achieves better overall performance than the pretrained 78B model from the same family on DriveBench, and surpasses GPT-5.1 on the planning dimension, providing insights toward efficient autonomous driving VLMs.
\end{abstract}

\begin{figure*}[t]
  \centering
  \includegraphics[width=\linewidth]{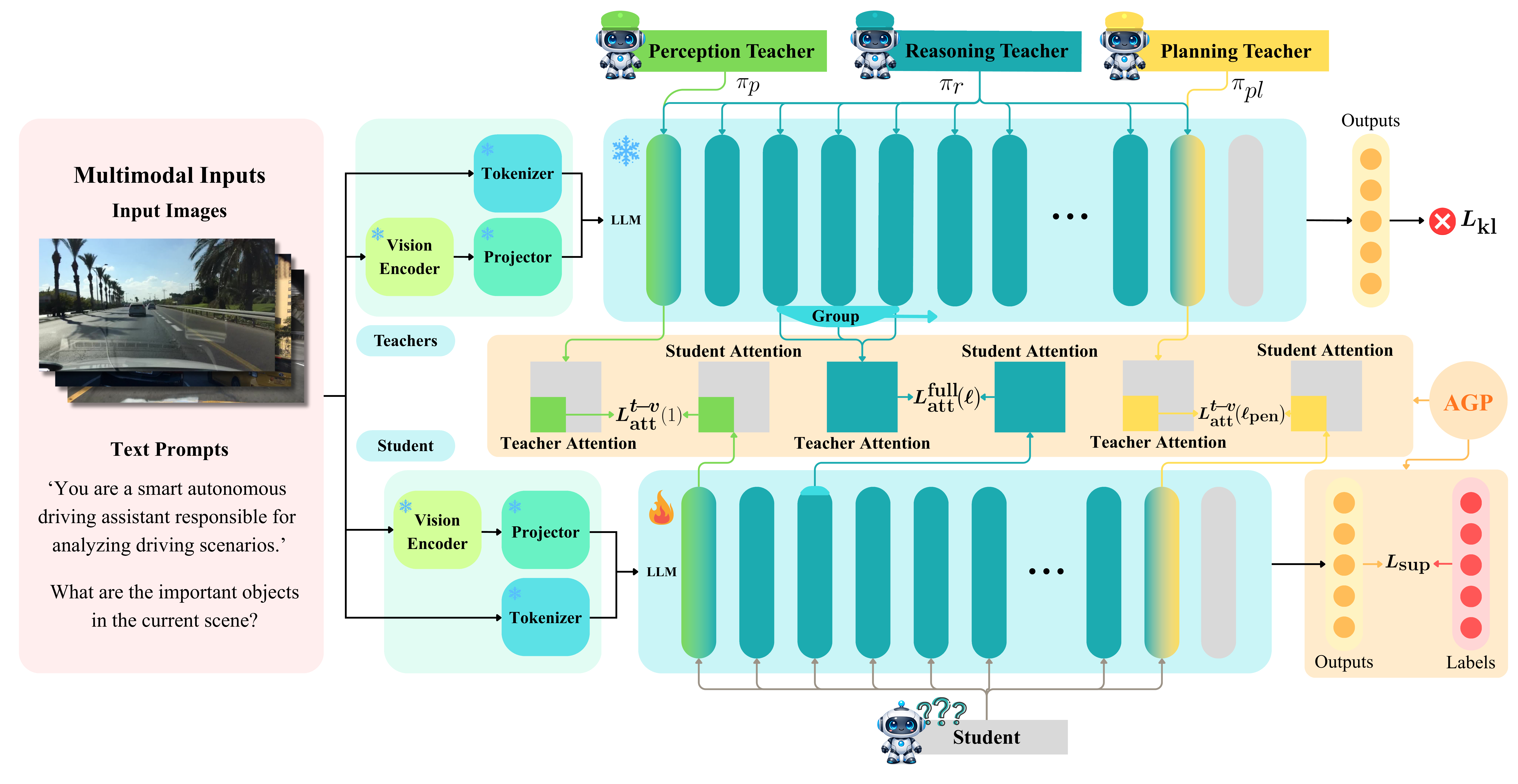}
  \caption{\textbf{Drive-KD multi-teacher distillation framework.} 
  Three teachers distill attention at different student layers: Perception/Planning uses first-layer/penultimate-layer cross-modal attention; Reasoning uses grouped-matching intermediate-layer full attention. The student is also trained with hard-label supervision at the output. We apply AGP to mitigate cross-capability gradient conflicts.}
  \label{fig:head}
\end{figure*}

\section{Introduction}
\label{sec:intro}
Autonomous driving is a highly complex domain with high safety requirements \cite{hu2023planning,zhao_autonomous_2024}.
Classical autonomous driving solutions mainly follow two paradigms \cite{yurtsever2020survey}.
One is a modular pipeline that decomposes object detection, tracking, behavior prediction, and trajectory planning into separate submodules \cite{vougioukas2019agricultural,paden2016survey,badue2021self}, and the other is an end-to-end learning paradigm that directly maps sensor inputs to control commands or semantic representations \cite{bojarski2016end,tampuu2020survey,chen2024end}.
Despite their progress, both paradigms have key limitations.
Modular pipelines often accumulate errors across stages, while end-to-end models remain hard to interpret and largely mimic driving behaviors without reasoning, leading to robustness and safety issues \cite{badue2021self,pan2024reliability,chen2024end}.

In recent years, vision-language models (VLMs), built on large language models (LLMs), demonstrate strong capabilities in zero-shot transfer and language-based reasoning, highlighting their potential for autonomous driving \cite{achiam2023gpt,wang2024qwen2,qwen2.5-VL,chen2024internvl}.
A number of recent studies have begun to adapt VLMs to autonomous driving, ranging from open-loop scene understanding to language-guided closed-loop driving systems, but they still face challenges in reliability and real-time performance \cite{tian_drivevlm_2024,sima_drivelm_2025,shao_lmdrive_2024}.

Knowledge distillation (KD) transfers knowledge from large models to small ones, enabling efficient deployment on resource-constrained edge devices \cite{xu_survey_2024,moslemi2024survey}.
However, current approaches lack systematic studies on distillation signal selection in autonomous driving, leaving limited guidance on where to apply supervision \cite{feng_align-kd_2024,yang_learning_2025}.
Moreover, many works treat driving capabilities as parallel objectives, ignoring their sequential dependencies \cite{tian_drivevlm_2024,shao_lmdrive_2024}.
When multiple distillation objectives are jointly optimized, a common practice is to aggregate them with a fixed or heuristic weighted sum, which can improve one capability at the cost of degrading others \cite{Pham_2023_WACV,Zhang_2023_MMKD,chen2024knowledge,zhong_revisiting_2024}.

Such trade-offs are unacceptable in autonomous driving, where perception, reasoning, and planning must be jointly reliable to ensure safe decision-making.

To address these limitations, we propose Drive-KD (\cref{fig:head}), a multi-teacher distillation framework for small-scale VLMs, aimed at improving their capability on comprehensive driving VQA tasks to better support high-level language guidance in autonomous driving.
We decompose autonomous driving into a sequential triad of capabilities, ``perception--reasoning--planning'', where later stages build on earlier-stage information.
Based on this capability decomposition, we first investigate layer selection strategies from both representation and capability perspectives.
We then identify suitable supervision signals for capability transfer by comparing their dispersion, and analyze why output-distribution alignment is not suitable for autonomous driving tasks.

Based on these findings, we systematically study distillation recipes for the three capabilities in autonomous driving, achieving capability scores that significantly outperform both pretrained and SFT baselines.
These results further validate the layer and signal selection principles identified in our initial analyses.
Moreover, we build Drive-KD as a multi-teacher distillation framework over the three capabilities, and apply asymmetric gradient projection (AGP) to mitigate gradient conflicts, yielding stronger capabilities in the small model.
We further demonstrate the generality of Drive-KD across model scales and different model families.

We summarize our contributions as follows:
\begin{itemize}
    \item First, we conduct the first systematic study of distillation design for autonomous driving VLMs, providing choices for distillation layers and supervision signals, and highlighting the limitations of output-distribution alignment for autonomous driving.

    \item Second, we derive the single-teacher distillation recipes for each capability and show stable improvements over pretrained and SFT baselines, which further validates our distillation signal selection principles.

    \item Third, we develop Drive-KD as a multi-teacher distillation framework and introduce AGP to alleviate cross-capability gradient conflicts. We further verify the generality of Drive-KD across model scales and different model families.

    \item Fourth, we demonstrate that our distilled InternVL3-1B achieves strong efficiency--performance trade-offs---requiring $\sim$42$\times$ less GPU memory and delivering $\sim$11.4$\times$ higher throughput---while outperforming the pretrained 78B model from the same family on DriveBench and surpassing GPT-5.1 on planning.
    
\end{itemize}

\section{Related Work}
\label{sec:related}
\subsection{LLM / VLM in Autonomous Driving}
\label{sec:ai_in_autodri}
The rapid progress of LLMs has accelerated the development of multimodal models \cite{zhao2023survey,zhang2024vision}. 
Building on this trend, VLMs have emerged and show strong general capabilities in complex scene description, visual question answering, and multi-step reasoning \cite{lu2025internvl,zhu2024llava,wang2024qwen2}, which naturally match the requirements of autonomous driving \cite{zhou2024vision,you2026v2x,guo2024vlm,li2025spacedrive}.
Recent efforts explore VLM-based scene description and hierarchical planning, employ structured VQA for reasoning, and develop end-to-end driving systems with language-grounded low-level controls 
\cite{tian_drivevlm_2024,sima_drivelm_2025,shao_lmdrive_2024,xu_drivegpt4_2024,xu2025drivegpt4,hwang_emma_2025}.
However, current VLMs still suffer from issues such as hallucination, unstable decision-making, and poor real-time performance, which limit their effectiveness in higher-level language-guided driving systems \cite{guan2024hallusionbench,li2025benchmark,tang2026drivep2dprogressiveperceptiontodecisionbenchmark}.

\subsection{Distillation for LLM / VLM}
\label{sec:distill_for_ai}
KD aims to transfer knowledge from a large teacher model to a smaller student model, improving the student's performance under limited capacity \cite{xu2024survey,gou2021knowledge}.
It was originally dominated by output distribution alignment \cite{hinton2015distilling}.
In the era of LLMs, KD has gradually expanded to richer forms of supervision beyond outputs \cite{li2023distilling,wang2024layerwised,pan2021meta}.
For VLMs, distillation needs to compress not only language modeling ability but also transfer cross-modal alignment and fusion \cite{feng_align-kd_2024,gu_multi-mllm_2025,li_ammkd_2025}.
Accordingly, recent works combine output-level alignment with internal-signal distillation and adopt techniques such as multi-teacher distillation to convey finer-grained information \cite{li2023distilling,wang2024layerwised}. 
These VLM-oriented distillation methods have been widely applied to VQA compression and multimodal instruction following \cite{tian_low-rank_2025,cao2025move,neo_towards_2025}.
However, principled strategies for choosing distillation signals and coordinating distillation across tasks remain unclear \cite{liang2023less,wang2024layerwised,mansourian2025comprehensive}.

\section{Methodology}
\label{sec:method}
\begin{figure*}[t]
  \centering
  \includegraphics[width=\textwidth]{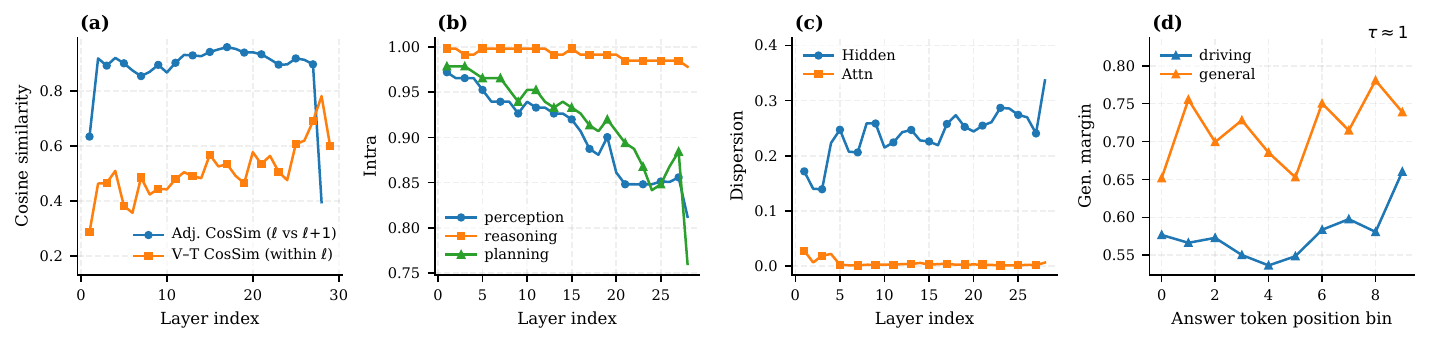}
  \caption{Pre-study summary for InternVL3-8B: (a) layer-wise distillation alignment measured by cosine similarity (adjacent-layer and within-layer vision--text), (b) capability-wise intra-group similarity across layers, (c) layer-wise dispersion of hidden states and attention maps ($1-\cos$), and (d) position-normalized generalized margin along the answer segment comparing driving and general data at $\tau\!\approx\!1.0$.}

  \label{fig:prestudy_all_internvl3_8b_combined}
\end{figure*}
\subsection{Preliminary}
\label{sec:preliminary}
We categorize driving capabilities into three sequential capabilities: perception, reasoning, and planning.
Perception aims to identify semantic cues that form the basis for subsequent reasoning.
Reasoning leverages these cues to infer relations and constraints among the environment.
Planning then leverages the inferred information to determine safe driving actions.
We aim to distill the three capabilities from a large teacher to a small student VLM.
Appendix~\ref{app:capability_decomposition} provides supporting evidence for this capability organization.

\subsection{Distillation Signal Selection}
\label{sec:prestudies}
We investigate (i) the appropriate layers for distilling each capability and (ii) the specific signals to distill within the selected layers. In addition, we revisit output-distribution alignment and analyze its limitations for autonomous driving. In our studies, we sample 1{,}000 images, and each image contains multiple questions under each capability.
\cref{fig:prestudy_all_internvl3_8b_combined} summarizes these analyses and supports our algorithm design.
Appendix~\ref{app:prestudy} collects the extended pre-study analyses: layer-wise similarity in Appendix~\ref{app:prestudy_similarity}, intra-consistency in Appendix~\ref{app:prestudy_intra}, hidden-state versus attention dispersion in Appendix~\ref{app:prestudy_dispersion}, and KL analyses in Appendix~\ref{app:prestudy_kl}. 

\subsubsection{Where To Distill: Layer Selection}
\label{sec:pre_where}
Distilling all layers is costly and often unnecessary: different layers encode different functions, and forcing the student to match the teacher everywhere can introduce redundant or even conflicting supervision \cite{liang2023less}.
Instead, we select a small set of target layers that are most informative for each driving capability.
We select distillation layers from two complementary views.
From a representation view, we pick layers with strong representation change and clear cross-modal interaction.
From a capability view, we use capability-wise intra-consistency to find layers with stable capability features across images.

\noindent\textbf{Representation view.}
We use two layer-wise metrics: adjacent-layer cosine similarity of hidden states and within-layer vision--text cosine similarity.
These metrics are computed on the sampled pre-study set.
For Transformer layers $\ell=1,\dots,N$, let hidden states be $\mathbf{H}^{(\ell)}\in\mathbb{R}^{B\times L\times D}$, with token vectors $\mathbf{h}^{(\ell)}_{b,i}\in\mathbb{R}^{D}$ at position $i$ of sample $b$.
Treating the input embeddings as layer $\ell=0$, we write $\mathbf{H}^{(0)}\in\mathbb{R}^{B\times L\times D}$ with token vectors $\mathbf{h}^{(0)}_{b,i}$, and the penultimate layer index is $\ell_{\mathrm{pen}}=N-1$.
We define adjacent-layer similarity as cosine similarity at matched token positions, averaged over samples and token indices:
\begin{equation}
S^{(\ell)}=\mathbb{E}_{b,i}\!\left[
\frac{\mathbf{h}^{(\ell)}_{b,i}\cdot \mathbf{h}^{(\ell+1)}_{b,i}}
{\|\mathbf{h}^{(\ell)}_{b,i}\|_2\;\|\mathbf{h}^{(\ell+1)}_{b,i}\|_2}
\right], \qquad \ell=0,\dots,N-1.
\end{equation}
To measure intra-layer vision--text similarity, we split tokens into vision and text segments and mean-pool each segment:
\begin{equation}
\bar{\mathbf{v}}^{(\ell)}_b=\frac{1}{|\mathcal{V}_b|}\sum_{i\in\mathcal{V}_b}\mathbf{h}^{(\ell)}_{b,i},
\qquad
\bar{\mathbf{t}}^{(\ell)}_b=\frac{1}{|\mathcal{T}_b|}\sum_{j\in\mathcal{T}_b}\mathbf{h}^{(\ell)}_{b,j}.
\end{equation}
Here $\mathcal{V}_b$ and $\mathcal{T}_b$ denote the index sets of visual and text tokens for sample $b$.
We then compute cosine similarity:
\begin{equation}
C^{(\ell)}=\mathbb{E}_{b}\!\left[
\frac{\bar{\mathbf{v}}^{(\ell)}_b\cdot \bar{\mathbf{t}}^{(\ell)}_b}
{\|\bar{\mathbf{v}}^{(\ell)}_b\|_2\;\|\bar{\mathbf{t}}^{(\ell)}_b\|_2}
\right].
\end{equation}
Across all models, $S^{(0)}$ is low while $S^{(1)}$ rises sharply, and $S^{(N-1)}$ drops relative to $S^{(N-2)}$, highlighting major representation changes near Layer~$1$ and $\ell_{\mathrm{pen}}$.
Meanwhile, $C^{(\ell)}$ exhibits a clear fusion peak near the penultimate layer ($\ell_{\mathrm{pen}}$), indicating strong vision--text interaction.
These results suggest that Layer $1$ and Layer $\ell_{\mathrm{pen}}$ serve as key informative layers, and we treat them as primary candidates.\\

\noindent\textbf{Capability view.}
The representation view is insufficient to identify layers that encode stable, capability-level structure across images, so we further use capability-wise intra-consistency as a complementary criterion for layer selection.
On the sampled dataset, for each layer $\ell$ and capability $c$, we compute an image--capability center by averaging the hidden vectors of the last non-special token over all questions of capability $c$ for the same image:
\begin{equation}
\mathbf{c}_{c,\mathrm{img}_g}^{(\ell)}=\frac{1}{N_{c,g}}\sum_{j=1}^{N_{c,g}}\mathbf{h}_{c,g,j}^{(\ell)},
\end{equation}
where $\mathbf{h}_{c,g,j}^{(\ell)}$ is the layer-$\ell$ hidden state of the last non-special token for the $j$-th question of capability $c$ on image $g$.
Let $\{\mathbf{c}_{c,\mathrm{img}_1}^{(\ell)},\dots,\mathbf{c}_{c,\mathrm{img}_G}^{(\ell)}\}$ be the centers of capability $c$ across $G$ images; we define intra as the mean pairwise cosine similarity:
\begin{equation}
\mathrm{intra}_{c}^{(\ell)}=\frac{2}{G(G-1)}\sum_{i<j}\cos\!\left(\mathbf{c}_{c,\mathrm{img}_i}^{(\ell)},\mathbf{c}_{c,\mathrm{img}_j}^{(\ell)}\right),
\end{equation}
where $\cos(\mathbf{a},\mathbf{b})=\left\langle \mathbf{a}/\|\mathbf{a}\|_2,\mathbf{b}/\|\mathbf{b}\|_2 \right\rangle$ and the summation is over all unordered image pairs ($i<j$).

Across all models, we observe consistent capability-specific patterns: perception has the highest intra at Layer $1$ and then decreases with depth, reasoning keeps a high intra across layers, and planning decreases overall but shows a clear peak near Layer $\ell_{\mathrm{pen}}$; together with the representation view, these results support selecting Layer $1$ for perception, using broader intermediate layers for reasoning to connect early perception cues with later decision features, and using the penultimate layer for planning to leverage late-stage capability specialization while retaining upstream information.

\subsubsection{What to Distill: Signal Selection}
\label{sec:pre_what}
For each (image, question) instance, we extract (i) the last non-special-token hidden vector and (ii) the head-mean attention matrix at the target layer, and measure dispersion across questions for the same image using $1-\cos(\cdot,\cdot)$, averaged over all question pairs and then over images.

For hidden states, for any two questions $i$ and $j$ paired under the same image, we compute the dispersion as
$D_{\mathrm{hid}}(i,j)=1-\cos(u_i,u_j)$, where $u_i$ (resp.\ $u_j$) is the hidden vector at the target layer for the last token of the generated answer segment excluding special tokens.

For attention, we use the head-mean attention map $\bar A$ and define
$D_{\mathrm{att}}(i,j)=1-\cos(\mathrm{vec}(\tilde A_i), \mathrm{vec}(\tilde A_j))$,
where $\tilde A_i$ and $\tilde A_j$ are the overlap-aligned top-left submatrices of $\bar A_i$ and $\bar A_j$ with size $m=\min(L_i,L_j)$, corresponding to the shared prefix tokens (vision tokens + system prompt), and are flattened by $\mathrm{vec}(\cdot)$ before cosine.

Across all three capabilities and all models, attention shows consistently lower dispersion than hidden states, indicating that attention reflects capability-stable behaviors more stably; therefore, we use attention as the distillation signal.

\subsubsection{Revisiting Output-Distribution Alignment}
\label{sec:exp_kl}
Output-distribution alignment is implemented as KL divergence between the teacher and student predictive distributions.
However, its effectiveness is highly sensitive to teacher capability and task complexity.
Across multiple temperatures, we feed the teacher model with both autonomous driving data and generic multimodal QA data~\cite{xu2024visionflanscalinghumanlabeledtasks}, and quantify its confidence on answer tokens using top-1 probability, top-10 probability mass, and the generalized margin.

Let teacher logits at answer position $t$ be $\mathbf{z}_t\in\mathbb{R}^{|\mathcal{V}|}$ and define the temperature-scaled softmax
$p_t^{(\tau)}(k)=\mathrm{softmax}\!\left(\mathbf{z}_t/\tau\right)_k$, we compute
\begin{equation}
m_t^{(\tau)} = \max_k\, p_t^{(\tau)}(k),
\end{equation}
\begin{equation}
S_t^{(\tau)} = \sum_{i=1}^{10} p_t^{(\tau)}(k_i),
\end{equation}
\begin{equation}
\Delta_{t,\mathrm{gen}}^{(\tau)} = m_t^{(\tau)} - \frac{S_t^{(\tau)}-m_t^{(\tau)}}{10-1}.
\end{equation}
where $k_1,\dots,k_{10}$ index the top-10 tokens under $p_t^{(\tau)}$.
The generalized margin $\Delta_{t,\mathrm{gen}}^{(\tau)}$ measures the dominance of the top-1 token over the other top-10 candidates; larger values indicate higher confidence.
Across all temperatures, driving outputs are consistently less confident and more diffuse than general outputs, making output-distribution alignment a noisy distillation signal; therefore, we do not include it as a distillation loss in Drive-KD.

\subsection{Single-Teacher Distillation}
\label{sec:single}
\subsubsection{Distillation Signals and Objectives}
\label{sec:single_obj}
We optimize the student with the hard-label supervised loss on our driving distillation dataset and add capability-specific attention distillation terms from a teacher.

\noindent\textbf{Hard-label supervised loss.}
Given a token sequence $\{y_t\}_{t\in \mathcal{Y}_{\mathrm{ans}}}$ over the answer span $\mathcal{Y}_{\mathrm{ans}}$, the hard-label supervised loss is
\begin{equation}
L_{\mathrm{sup}}  = - \frac{1}{|\mathcal{Y}_{\mathrm{ans}}|}\sum_{t\in \mathcal{Y}_{\mathrm{ans}}}\log p_\theta(y_t \mid x, y_{<t}).
\end{equation}
\noindent\textbf{Attention Maps.}
Let $A^{(\ell)}_{s},A^{(\ell)}_{t}\in \mathbb{R}^{B \times H \times L \times L}$ denote the attention weights at layer $\ell$ for the student and teacher.
We first average attention over heads, where $(\cdot)\in\{s,t\}$:
\begin{equation}
\bar A^{(\ell)}_{(\cdot),b,i,j}
= \frac{1}{H}\sum_{h=1}^{H} A^{(\ell)}_{(\cdot),b,h,i,j}
\end{equation}

\noindent\textbf{Full-attention distillation.}
We distill the full attention matrix by matching all query-to-key entries on valid positions between the student and teacher at layer $\ell$, where $i$ indexes the query token position and $j$ indexes the key token position in a length-$L$ sequence.
\begin{equation}
L_{\mathrm{att}}^{\mathrm{full}}(\ell)
= \operatorname{mean}_{b,i}\!\left[
\operatorname{mean}_{j}\!\left(\bar A^{(\ell)}_{s,b,i,j}-\bar A^{(\ell)}_{t,b,i,j}\right)^2
\right].
\end{equation}
We compute the squared error for each query-token $(b,i)$, average it over keys $j$, and then average over $(b,i)$.\\

\noindent\textbf{Text-to-vision attention distillation.}
We compute the loss only over text query positions and normalize within each sample by the number of text-to-vision pairs.
Let $T_b$ and $V_b$ denote the sets of text-query positions and vision-key positions in sample $b$, respectively, where $T_b$ includes all text tokens in the prompt+answer sequence excluding special and padding tokens.
The corresponding distillation loss is
\begin{equation}
\begin{aligned}
L_{\mathrm{att}}^{t\!-\!v}(\ell)
&= \operatorname{mean}_{b}\;
\frac{1}{|T_b|\,|V_b|}
\sum_{i\in T_b}\sum_{j\in V_b}
\left(\bar A^{(\ell)}_{s,b,i,j}-\bar A^{(\ell)}_{t,b,i,j}\right)^2 .
\end{aligned}
\end{equation}

We apply an online loss reweighting based on per-loss gradient magnitudes to stabilize multi-loss training. Further implementation details are provided in Appendix~\ref{online_loss}. 
In this single-teacher stage, we do not use gradient projection, as it would dampen the attention-distillation gradients and thus confound a clean comparison among different distillation signals.
\subsubsection{Capability-Specific Recipes}
\label{sec:single_recipe}
We use hard-label supervision for all capabilities, and add capability-specific attention distillation as the soft term.
For perception, we distill Layer~$1$ text-to-vision attention, $A_{1,t\!-\!v}$.
For reasoning, we distill intermediate-layer attention with layer-group matching \cite{kim_compodistill_2025}.
Let the selected student layers be $\ell_1,\dots,\ell_K$ and the selected teacher layers be $k_1,\dots,k_M$, where both sets are restricted to the range from Layer $1$ to the penultimate layer $\ell_{\mathrm{pen}}$; we set group size $g=\max(1,\,M-K+1)$ and assign each student layer $\ell_i$ to a consecutive sliding teacher group $\mathcal{G}(\ell_i)=\{k_{s_i+1},\dots,k_{s_i+g}\}$ with $s_i=\min(i-1,\,M-g)$, using the mean attention over $\mathcal{G}(\ell_i)$ as the target.
For planning, we distill penultimate-layer text-to-vision attention, $A_{\ell_{\mathrm{pen}},t\!-\!v}$.

\subsection{Multi-Teacher Distillation}
\label{sec:multi}

\subsubsection{Objective and Teacher Mixing}
\label{sec:multi_obj}
Following the single-teacher recipes in \cref{sec:single_recipe}, we define Drive-KD as a multi-teacher distillation framework with three teachers specialized for each capability, and we apply all three teachers to every batch.
For a batch from capability $c$, we combine the three teachers with a fixed weight $\pi_{c,t}$, which sets how much teacher $t$ contributes on capability $c$ (see Appendix~\ref{app:teacher_mixing}).
Each teacher provides one capability-specific attention distillation loss $L_t$.

We use the following three definitions:
\begin{equation}
L_p \;=\; L_{\mathrm{att}}^{t\!-\!v}(1).
\end{equation}
\begin{equation}
L_r
= \frac{1}{|\mathcal{S}|}\sum_{\ell\in \mathcal{S}}
\operatorname{mean}_{b,i}\!\left[
\operatorname{mean}_{j}\!\left(
\bar A^{(\ell)}_{s,b,i,j}
- \mu^{(\ell)}_{t,b,i,j}
\right)^2
\right].
\end{equation}
\begin{equation}
\mu^{(\ell)}_{t,b,i,j}
= \operatorname{mean}_{k\in \mathcal{G}(\ell)} \bar A^{(k)}_{t,b,i,j}.
\end{equation}
\begin{equation}
L_{pl} \;=\; L_{\mathrm{att}}^{t\!-\!v}(\ell_{\mathrm{pen}}), 
\quad \ell_{\mathrm{pen}} = N_s-1,
\end{equation}
Here $\mathcal{S}$ is the selected student layer set, and $\mathcal{G}(\ell)$ maps each $\ell\in\mathcal{S}$ to a small group of nearby teacher layers.

With these definitions, the total loss combines hard-label supervision with mixed distillation:
\begin{equation}
L = w_{\mathrm{ce}}L_{\mathrm{sup}}  + \sum_{t\in\{p,r,pl\}} \pi_{c,t}\, w_t\, L_{t}.
\end{equation}
\subsubsection{Asymmetric gradient projection}
\label{sec:agradient_projection}

To reduce gradient conflicts, we adopt AGP with two stages.

\begin{figure*}[t]
  \centering
  \includegraphics[width=1.0\textwidth]{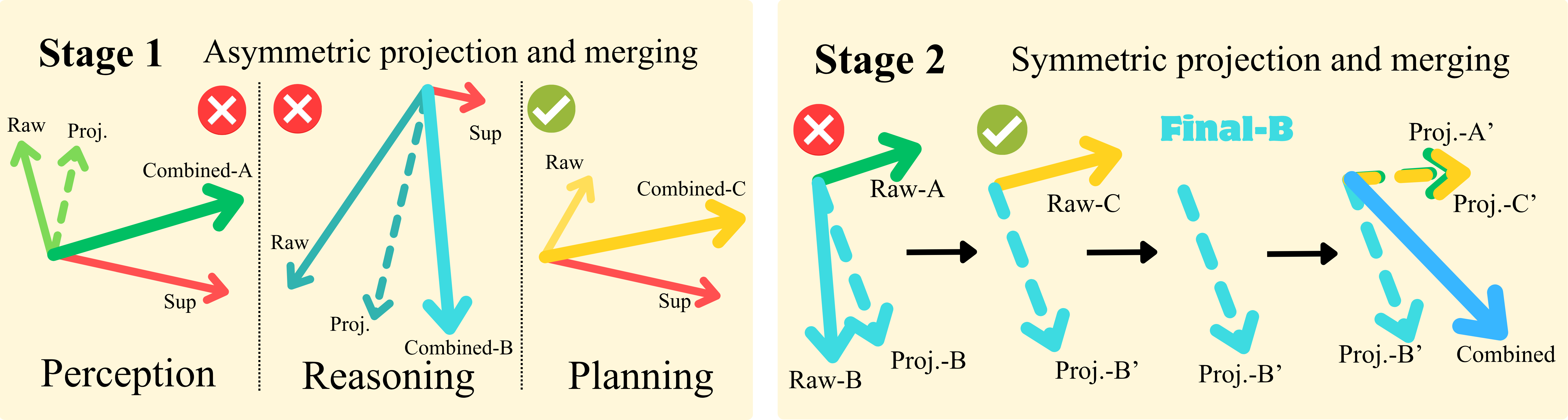}
  \caption{\textbf{Asymmetric Gradient Projection (AGP).}
  Stage~1 uses an asymmetric anchor--follower projection within each capability and merges the update.
  Stage~2 applies shuffled symmetric pairwise projections across capabilities to obtain the final gradient direction (gradient~B shown as an example).}

  \label{fig:agp}
\end{figure*}

\noindent\textbf{Stage 1: Asymmetric capability-wise projection and merging.}
For capability $c$, define
$\mathbf{g}_{\mathrm{sup}}=\nabla_\theta\!\left(w_{\mathrm{ce}}L_{\mathrm{sup}}\right)$
and
$\mathbf{g}_{c}=\nabla_\theta\!\left(\sum_{t\in\{p,r,pl\}} \pi_{c,t}\, w_t\, L_t\right)$;
we use $\mathbf{g}_{\mathrm{sup}}$ as anchor and $\mathbf{g}_c$ as follower.
\begin{equation}
\mathbf{g}_{a}^{(c)} = \mathbf{g}_{\mathrm{sup}},\qquad
\mathbf{g}_{f}=\mathbf{g}_{c}.
\end{equation}
Then, we remove only the conflicting component of the follower along the anchor direction, while keeping the anchor unchanged:
\begin{equation}
\mathbf{g}_f \leftarrow \mathbf{g}_f - \frac{\mathbf{g}_f^\top \mathbf{g}_a^{(c)}}{\|\mathbf{g}_a^{(c)}\|_2^2}\,\mathbf{g}_a^{(c)},\quad
\text{if }\mathbf{g}_f^\top \mathbf{g}_a^{(c)}<0
\end{equation}
We then merge the anchor and the follower to obtain the capability-level gradient used in Stage~2:
\begin{equation}
\mathbf{g}^{(c)}=\mathbf{g}_a^{(c)}+\mathbf{g}_f.
\end{equation}

\noindent\textbf{Stage 2: Symmetric pairwise projections among merged capability gradients.}
We have three capabilities $\mathcal{C}=\{p,r,pl\}$ and obtain merged gradients
$\mathbf{g}^{(p)},\mathbf{g}^{(r)},\mathbf{g}^{(pl)}$ from Stage~1.
To mitigate cross-capability conflicts, we check conflicts in a shuffled pairwise order and apply a projection only when a conflict is detected.
\begin{equation}
\tilde{\mathbf{g}}^{(c_1)} \leftarrow \mathbf{g}^{(c_1)},\quad
\tilde{\mathbf{g}}^{(c_2)} \leftarrow \mathbf{g}^{(c_2)},\quad
\tilde{\mathbf{g}}^{(c_3)} \leftarrow \mathbf{g}^{(c_3)}.
\end{equation}
Specifically, we sample a random permutation $(c_1,c_2,c_3)$ of $(p,r,pl)$ and sequentially update each $\tilde{\mathbf{g}}^{(c_i)}$ by projecting it against the other two gradients in the order $c_1\rightarrow c_2\rightarrow c_3$.
\begin{equation}
\tilde{\mathbf{g}}^{(c_i)} \leftarrow
\tilde{\mathbf{g}}^{(c_i)} -
\frac{\tilde{\mathbf{g}}^{(c_i)\top}\mathbf{g}^{(c_j)}}{\|\mathbf{g}^{(c_j)}\|_2^2}\,\mathbf{g}^{(c_j)},\quad
\text{if } \tilde{\mathbf{g}}^{(c_i)\top}\mathbf{g}^{(c_j)} < 0
\end{equation}
Finally, we sum the projected capability gradients to form the overall update direction:
\begin{equation}
\mathbf{g}=\tilde{\mathbf{g}}^{(p)}+\tilde{\mathbf{g}}^{(r)}+\tilde{\mathbf{g}}^{(pl)}.
\end{equation}

As shown in \cref{fig:agp}, Stage~1 uses the hard-label supervised gradient as an anchor and removes only the part of the distillation update that conflicts with it, avoiding sacrificing hard-label learning to fit weaker teachers' noisier or mismatched soft signals. Stage~2 resolves negative inner products among capability-level merged gradients to reduce the chance of improving one capability at the expense of another; the shuffled projection order further helps avoid consistently biasing toward any fixed capability.

\begin{table*}[t]
  \caption{Capability performance and deployment-oriented inference efficiency of representative VLMs on DriveBench. We report capability scores (\%) and inference metrics, including peak GPU memory (GB) under the minimum viable number of GPUs, average GPU-side generation throughput (tokens/s), and median GPU-side time-to-first-token (s).}
  \label{tab:vlm_capability_efficiency}
  \centering
  \begin{small}
    \setlength{\tabcolsep}{5pt}
    \begin{tabular}{lcccccccc}
      \toprule
      \multirow{2}{*}{Model} &
      \multicolumn{4}{c}{Capability scores (\%)} &
      \multicolumn{3}{c}{Deployment metrics} \\
      \cmidrule(lr){2-5}\cmidrule(lr){6-8}
      & Perception & Reasoning & Planning & Avg. & Memory (GB) & Speed (tok/s) & 1st-token (s) \\
      \midrule
      GPT-5.1 & 45.56 & 41.02 & 51.94 & \underline{\textbf{46.17}} & -     & -    & -    \\
      \midrule
      InternVL3-1B  & 33.26 & 20.96 & 22.36 & 25.53 & \underline{\textbf{4.1}} & \underline{\textbf{45.7}} & \underline{\textbf{0.45}} \\
      InternVL3-2B  & 37.71 & 35.99 & 26.19 & 33.30 & 6.3   & 39.9 & 0.67 \\
      InternVL3-8B  & 40.05 & 41.15 & 32.77 & 37.99 & 18.3  & 32.6 & 1.58 \\
      InternVL3-14B & 39.83 & 40.84 & 36.35 & 39.01 & 33.4  & 17.0 & 3.01 \\
      InternVL3-38B & 34.27 & 38.48 & 40.65 & 37.80 & 87.0  & 7.3  & 9.84 \\
      InternVL3-78B & 42.01 & \underline{\textbf{47.16}} & 36.31 & 41.83 & 171.6 & 4.0  & 16.46 \\
      \midrule
      Qwen2.5-VL-3B-Instruct  & 35.46 & 30.81 & 25.29 & 30.52 & 8.5 & 28.0 & 0.68 \\
      Qwen2.5-VL-3B-DriveLM & 30.19 & 35.28 & 35.41 & 33.63 & 8.5 & 28.0 & 0.68 \\
      Qwen2.5-VL-7B-Instruct  & 36.26 & 37.54 & 32.84 & 35.55 & 17.1  & 32.0 & 0.87 \\
      Qwen2.5-VL-32B-Instruct & 38.41 & 41.30 & 34.29 & 38.00 & 69.5  & 10.8 & 2.36 \\
      Qwen2.5-VL-72B-Instruct & 23.78 & 27.67 & 50.76 & 34.07 & 146.5 & 5.8  & 4.26 \\
      \midrule
      Llama-3.2-11B-Vision-Instruct & 31.59 & 32.91 & 29.34 & 31.28 & 26.2  & 16.8 & 1.55 \\
      Llama-3.2-90B-Vision-Instruct & 27.26 & 26.33 & 27.72 & 27.10 & 183.6 & 2.7  & 8.05 \\
      \midrule
      InternVL3-1B (Single)           & 43.13 & 34.32 & 52.97 & 43.47 & \underline{\textbf{4.1}} & \underline{\textbf{45.7}} & \underline{\textbf{0.45}}  \\
      Qwen2.5-VL-3B-Instruct (Single) & 45.59 & 34.47 & 51.97 & 44.01 & 8.5 & 28.0 & 0.68  \\
      \midrule
      InternVL3-1B (Multi)            & 43.50 & 33.15 & \underline{\textbf{55.51}} & 44.05 & \underline{\textbf{4.1}} & \underline{\textbf{45.7}} & \underline{\textbf{0.45}} \\
      Qwen2.5-VL-3B-Instruct (Multi)  & \underline{\textbf{45.63}} & 36.41 & 54.07 & 45.37 & 8.5 & 28.0 & 0.68 \\  
      \bottomrule
    \end{tabular}
  \end{small}
\end{table*}

\section{Experiments}
\label{sec:exp}

\subsection{Experimental Setup}
\label{sec:exp_setup}

\noindent\textbf{Data and prompts.}
We train on a 10,500 human-annotated driving distillation dataset with single- and multi-view VQA; Appendix~\ref{app:data_diversity} reports diversity statistics and Appendix~\ref{app:annotations} provides examples.

\noindent\textbf{Models and training.}
Our main experiments consider two teacher--student pairs: InternVL3-8B $\rightarrow$ InternVL3-1B and Qwen2.5-VL-7B-Instruct $\rightarrow$ Qwen2.5-VL-3B-Instruct.
We provide reproducibility details in Appendix~\ref{app:reproducibility}.

\noindent\textbf{Evaluation.}
Evaluation is conducted on DriveBench \cite{xie2025drivebench}. 
We merge DriveBench's original prediction and behavior dimensions and report them jointly as reasoning.
We use DeepSeek-V3.2 as the evaluator \cite{liu2025deepseek}.
We report capability scores for perception, reasoning, and planning, together with the overall average, and deployment-oriented inference metrics.
Additional training and evaluation details, including evaluation-case illustrations, are provided in Appendix~\ref{app:train_eval}.
Additional AutoDriDM~\cite{tang2026drivep2dprogressiveperceptiontodecisionbenchmark} results and VLM distillation baselines are in Appendix~\ref{app:addition}. 

\subsection{Single-Teacher Evaluation}
\label{sec:exp_single}
We evaluate single-teacher distillation by distilling each student from a single teacher within its model family. Each capability-specific single-teacher student is trained only on the corresponding capability split of our distillation dataset (Appendix~\ref{app:data_diversity}).
In our setting, CE-only training is equivalent to SFT; we compare against this baseline in \cref{sec:exp_ablation}.

As summarized in \cref{tab:vlm_capability_efficiency}, single-teacher distillation yields consistent gains over pretrained baselines across both student families. Concretely, InternVL3-1B improves from 25.53 to 43.47 in Avg.\ score, and Qwen2.5-VL-3B-Instruct improves from 30.52 to 44.01. On planning, the distilled students become competitive with GPT-5.1: InternVL3-1B (Single) reaches 52.97 and Qwen2.5-VL-3B-Instruct (Single) reaches 51.97, compared with 51.94 for GPT-5.1, and outperform the autonomous-driving VLM ReasonDrive~\cite{chahe2025reasondriveefficientvisualquestion} (Qwen2.5-VL-3B-DriveLM). These results show that our distillation can transfer driving-related behaviors to small students effectively.

Across these capabilities, the results follow a consistent trend. Perception improves strongly, matching our choice of early cross-modal attention for supervision. Reasoning also improves over pretrained baselines, but remains the hardest dimension: the gap to GPT-5.1 is larger than that in perception and planning, suggesting multi-step relation understanding is more difficult to transfer to a small VLM. Planning shows the largest gains and becomes the strongest dimension for distilled students, which supports our design of using late-stage cross-modal attention.

\subsection{Multi-Teacher Evaluation}
\label{sec:exp_multi}
Compared with the single-teacher, multi-teacher training achieves a higher overall score (see \cref{tab:vlm_capability_efficiency}). For InternVL3-1B, it improves Avg.\ from 43.47 to 44.05 and planning from 52.97 to 55.51, while preserving strong perception, highlighting the complementary benefits of Drive-KD.

By capability, perception stays on par with single-teacher, suggesting early-layer cross-modal supervision is already sufficient. Reasoning remains comparable but appears most sensitive to cross-capability interference under multi-objective training. Planning improves the most and drives the higher average score, consistent with its reliance on integrating perception cues with intermediate reasoning signals, which multi-teacher supervision strengthens. 

In terms of efficiency, the distilled models retain the lightweight deployment profile of small VLMs while achieving much stronger capability scores. For example, InternVL3-1B (Multi) uses 4.1\,GB peak memory with 45.7 tok/s throughput, being far more efficient than pretrained InternVL3-78B, yielding a better capability--efficiency trade-off.

In the overall results (\cref{tab:vlm_capability_efficiency}), GPT-5.1 scores higher on reasoning than the Drive-KD InternVL3-1B and Qwen2.5-VL-3B-Instruct students, but does not outperform them on planning.
This motivates testing whether higher reasoning necessarily translates to better planning on the same visual inputs.
For each image $k$, we average evaluator scores over its reasoning and planning questions to obtain $R_k$ and $P_k$:
\begin{equation}
R_k=\operatorname{mean}_{i\in\mathcal{Q}_r(k)} s_i,
\qquad
P_k=\operatorname{mean}_{j\in\mathcal{Q}_p(k)} s_j,
\end{equation}
where $s_i$ is the evaluator score for one (image, question) instance.
We then compute Pearson and Spearman correlations between $\{R_k\}$ and $\{P_k\}$ over images that contain both question types.
As shown in \cref{tab:rp_corr}, GPT-5.1 shows near-zero reasoning--planning correlation, whereas our students exhibit a weak positive association. This suggests that Drive-KD not only improves planning scores, but also strengthens same-scene reasoning--planning alignment, yielding a more consistent reasoning-to-planning mapping.

\begin{table}[t]
  \caption{Image-level correlation between reasoning and planning scores.}
  \label{tab:rp_corr}
  \centering
  \begin{small}
    \setlength{\tabcolsep}{6pt}
    \begin{tabular}{lcc}
      \toprule
      Model & Pearson($R_k$, $P_k$) & Spearman($R_k$, $P_k$) \\
      \midrule
      GPT-5.1 & $-0.0418$ & $-0.0316$ \\
      Ours (InternVL3) & $+0.3384$ & $+0.3724$ \\
      Ours (Qwen2.5-VL) & $+0.2976$ & $+0.3271$ \\
      \bottomrule
    \end{tabular}
  \end{small}
\end{table}

\subsection{Distillation Scaling Across Model Sizes}
\label{sec:exp_scaling}

\begin{table}[t]
  \caption{DriveBench scores (\%) of InternVL3 students distilled from InternVL3 teachers. Here we abbreviate model size by parameter scale (e.g., 1B/2B/8B/14B/38B).}
  \label{tab:teacher_student_size_ablation}
  \centering
  \begin{small}
    \begin{tabular}{cccccc}
      \toprule
      Tea & Stu & Perception & Reasoning & Planning & Avg. \\
      \midrule
      8B  & 1B & \underline{\textbf{43.50}} & 33.15 & 55.51 & 44.05 \\
      14B & 1B & 43.41 & 30.34 & 56.19 & 43.31 \\
      38B & 1B & 43.24 & 29.15 & 56.77 & 43.05 \\
      \midrule
      8B  & 2B & 43.14 & 36.97 & 56.01 & 45.37 \\
      14B & 2B & 41.74 & 35.40 & 56.84 & 44.66 \\
      38B & 2B & 42.87 & \underline{\textbf{38.25}} & \underline{\textbf{57.63}} & \underline{\textbf{46.25}} \\
      \bottomrule
    \end{tabular}
  \end{small}
\end{table}

We run the multi-teacher algorithm with teacher sizes $\{8\mathrm{B},14\mathrm{B},38\mathrm{B}\}$ and student sizes $\{1\mathrm{B},2\mathrm{B}\}$.
\cref{tab:teacher_student_size_ablation} shows that for a fixed student, larger teachers can improve, match, or degrade different capabilities, and the best teacher size depends on student capacity.
Meanwhile, under comparable distillation settings, the higher-capacity student is consistently stronger, indicating that student capacity remains crucial for absorbing transferred knowledge, and stronger teachers may be harder for small students to learn due to more complex feature representations. Overall, Drive-KD remains effective across different size pairings.

\begin{table}[t]
  \caption{InternVL3-1B ablations on DriveBench: distillation objectives and gradient-conflict handling.}
  \label{tab:ablation_distillation_signals}
  \centering
  \begin{small}
    \setlength{\tabcolsep}{4.5pt}
    \begin{tabular}{lccc}
      \toprule
      Setting & Perception & Reasoning & Planning \\
      \midrule
      CE                       & 40.86 & 29.05 & 45.63 \\
      CE + KL                  & 39.60 & 28.16 & 43.36 \\
      CE + Hidden (1)           & 41.27 & -     & -     \\
      CE + Hidden (mid)         & -     & 31.65 & -     \\
      CE + Hidden ($\ell_{\mathrm{pen}}$) & -     & -     & 45.04 \\
      Ours (single-teacher)    & \underline{\textbf{43.13}} & \underline{\textbf{34.32}} & \underline{\textbf{52.97}} \\
      \midrule
      Multi-teacher            & 42.34 & 25.68 & 51.03 \\
      Multi-teacher (G1)       & 42.96 & 25.49 & 46.99 \\
      Multi-teacher (G2)       & 42.64 & 29.18 & 52.19 \\
      Ours (multi-teacher)     & \underline{\textbf{43.50}} & \underline{\textbf{33.15}} & \underline{\textbf{55.51}} \\
      \bottomrule
    \end{tabular}
  \end{small}
\end{table}

\subsection{Ablation Study}
\label{sec:exp_ablation}
We ablate six aspects on InternVL3-1B: (i) whether to add output-distribution alignment (CE+KL) beyond hard-label CE, (ii) hidden-state distillation variants, and (iii) gradient-conflict handling in multi-teacher training (none, G1/G2, or AGP). Here (i) and (ii) follow the single-teacher protocol: the model is trained on the corresponding capability split only, and we report the score on that capability (Table~\ref{tab:ablation_distillation_signals}). Appendix~\ref{app:attn_variants} reports the remaining ablations in more detail, including per-capability layer choices in Appendix~\ref{app:layer_ablation}, alternative attention targets (full attention versus text-to-vision attention), and similarity-based attention matching.

CE is equivalent to SFT, and KL aligns the output token distribution.
``mid'' denotes the intermediate layers from the first to the penultimate; G1 applies PCGrad~\cite{NEURIPS2020_3fe78a8a} jointly across all losses, while G2 projects CE vs.\ soft within each capability and then projects across capabilities.

Overall, KL consistently degrades performance relative to CE, and comparisons against hidden-state distillation further support our signal choices; in contrast, our single-teacher recipe achieves the best overall performance.
In multi-teacher training, the plain baseline severely hurts reasoning, and joint PCGrad (G1) is also unstable---again most clearly on reasoning.
A two-stage symmetric scheme (G2) recovers reasoning and improves planning, but AGP achieves the best planning score while keeping perception and reasoning strong.
Together, these ablations support Drive-KD: capability-specific distillation signals and AGP are both critical for consistent gains.

\section{Conclusion}
\label{sec:concl}
We propose Drive-KD, a multi-teacher distillation framework for autonomous driving VLMs.
Drive-KD decomposes driving into a sequential triad of perception, reasoning, and planning, and transfers each capability with targeted internal supervision.
First, we present a systematic study of distillation design for driving VLMs.
We identify Layer~1 cross-modal attention as most informative for perception, broader intermediate layers for reasoning, and the penultimate layer for planning.
We find attention is more stable than hidden states, and output-distribution alignment is unreliable for autonomous driving.
Second, based on these findings, we derive strong single-teacher recipes for each capability and improve small VLMs over baselines.
Third, we unify the three capabilities into a multi-teacher framework and introduce AGP to reduce cross-capability interference.
Finally, experiments show that distilled models achieve strong capability--efficiency trade-offs and Drive-KD generalizes across model scales and families.
Our distilled InternVL3-1B uses $\sim$42$\times$ less peak GPU memory and achieves $\sim$11.4$\times$ higher generation throughput than pretrained InternVL3-78B, while outperforming it on DriveBench overall and surpassing GPT-5.1 on planning.
We hope Drive-KD enables efficient driving VLMs and better supports language-guided driving systems.

\section*{Impact Statement}

Drive-KD is expected to achieve faster on-vehicle response while improving driving capability. By lowering the hardware requirements for advanced AI safety functions, it may help make such functions available across a wider range of vehicle classes and price points. Compared with traditional end-to-end ``black-box'' models, our ``perception--reasoning--planning'' triad offers better interpretability.

Many recent language-grounded driving and vision-language-action (VLA) systems are fundamentally built upon vision-language models (VLMs). A significant portion of the challenges in closed-loop driving arises not only from control, but from limitations in upstream visual understanding, relational reasoning, and semantic decision-making. By improving the efficiency and capability of driving-oriented VLMs through structured distillation, our approach targets this critical upstream bottleneck and serves as a practical step toward more reliable closed-loop systems.

However, reasoning ability remains the most difficult aspect to transfer to smaller models. Importantly, our results are based on the open-loop evaluation of DriveBench and do not guarantee real-world performance. We strongly recommend that any distilled model produced by Drive-KD undergo rigorous closed-loop simulation and extensive long-tail and edge-case testing before deployment in any physical system. For safety-critical deployment, potential risks should be assessed with particular caution.

\section*{Limitations}

The current framework is limited to image-and-text inputs and does not incorporate other sensor modalities such as video or point clouds, resulting in underutilized temporal and geometric information. Transferring complex multi-step reasoning to smaller models remains a bottleneck, as reflected in the capability scores, and some high-level reasoning patterns are difficult to preserve under limited capacity. Our study focuses on VQA-based evaluation to assess VLM capabilities from a semantic perspective, emphasizing upstream exploration rather than downstream application.


\bibliography{ref}
\bibliographystyle{icml2026}

\newpage
\appendix
\onecolumn
\section{Capability Decomposition Evidence}
\label{app:capability_decomposition}

We organize driving capabilities into perception, reasoning, and planning, following the sequential dependency of driving decisions. 
Since planning ultimately determines driving actions, we examine its alignment with preceding capabilities, namely perception and reasoning. 
To examine this alignment, we train four InternVL3-1B SFT variants: planning only, perception+planning, reasoning+planning, and perception+reasoning+planning.
On 200 images, we aggregate image-level scores and compute Pearson and Spearman correlations between perception and planning, and between reasoning and planning. 
Table~\ref{tab:capability_decomposition_corr} reports the resulting image-level correlations under these capability-composition settings.

\begin{table}[H]
  \caption{Image-level correlations under different capability-composition SFT settings.}
  \label{tab:capability_decomposition_corr}
  \centering
  \begin{small}
    \begin{tabular}{lcccc}
      \toprule
      Setting & Pear.\ $(P,Pl)$ & Spear.\ $(P,Pl)$ & Pear.\ $(R,Pl)$ & Spear.\ $(R,Pl)$ \\
      \midrule
      Plan & +0.1599 & +0.1492 & -0.0498 & -0.0160 \\
      Perc.+Plan & +0.2234 & +0.2243 & -- & -- \\
      Reas.+Plan & -- & -- & +0.1144 & +0.0923 \\
      Perc.+Reas.+Plan & \underline{\textbf{+0.2516}} & \underline{\textbf{+0.2359}} & \underline{\textbf{+0.1712}} & \underline{\textbf{+0.1605}} \\
      \bottomrule
    \end{tabular}
  \end{small}
\end{table}

The correlations become stronger as preceding capabilities are included during training, suggesting that planning is more aligned with perception and reasoning when those supporting capabilities are learned jointly. 
This provides empirical evidence supporting the capability decomposition used in Drive-KD. 

\section{Additional Pre-study Analyses}
\label{app:prestudy}

\subsection{Layer-wise similarity profiles across model families and scales}
\label{app:prestudy_similarity}
\begin{figure}[H]
  \centering
  \includegraphics[width=\textwidth]{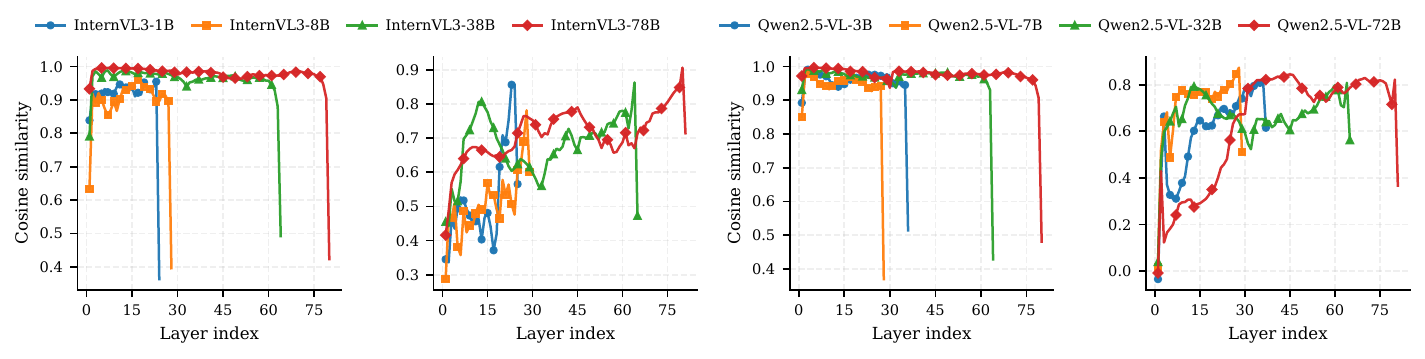}
  \caption{
    Layer-wise similarity profiles across model families and scales.
    Adjacent-layer cosine similarity (Adj. CosSim) and within-layer vision--text cosine similarity (V--T CosSim)
    are shown for InternVL3 (1B/8B/38B/78B) and Qwen2.5-VL (3B/7B/32B/72B).
  }
  \label{fig:appendix-similarity-allmodels}
\end{figure}

We report additional layer-wise similarity profiles (\cref{fig:appendix-similarity-allmodels}) to support our layer selection criteria across model families and parameter scales. 

\subsection{Capability-wise intra-consistency across Transformer layers}
\label{app:prestudy_intra}
\begin{figure}[!t]
  \centering
  \includegraphics[width=\textwidth]{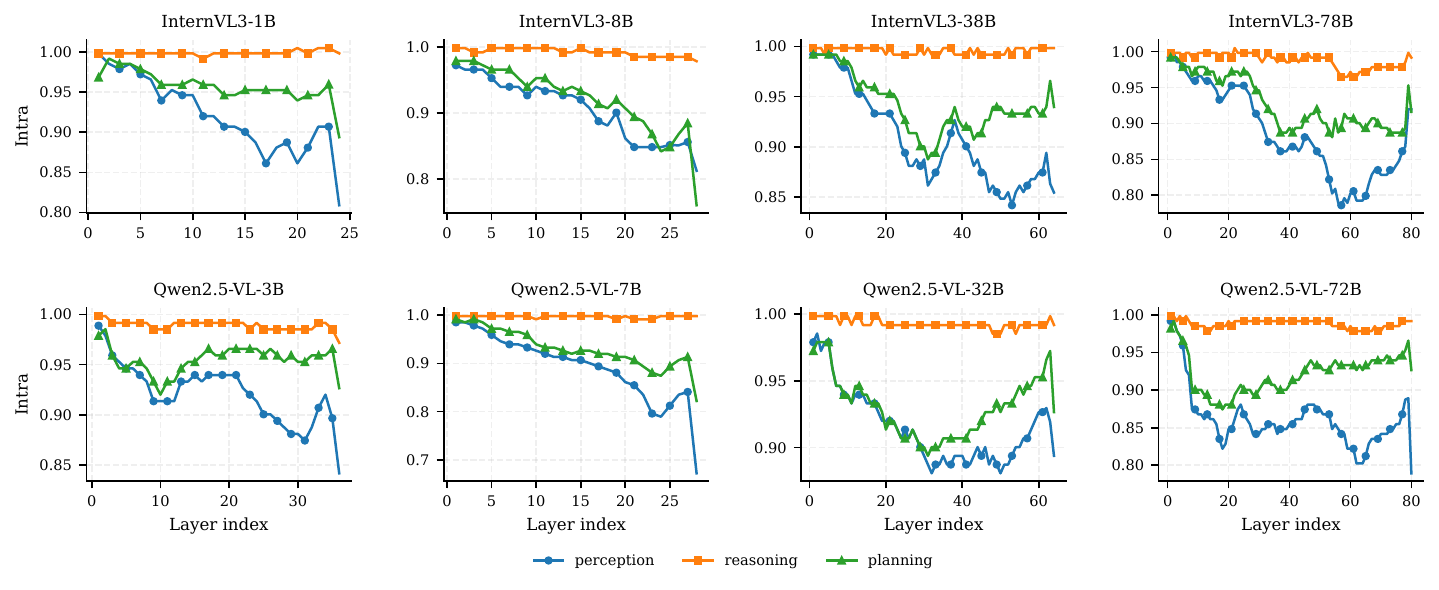}
  \caption{Capability-wise intra-consistency (mean pairwise cosine similarity) across Transformer layers for InternVL3 (top row) and Qwen2.5-VL (bottom row) under three task types: perception, reasoning, and planning.}
  \label{fig:appendix2-intra-all}
\end{figure}

We further include capability-wise intra-consistency curves (\cref{fig:appendix2-intra-all}), which complement representation-based signals by quantifying how stable each capability representation is across images at different depths. 

\subsection{Hidden-state vs.\ attention dispersion under rewording}
\label{app:prestudy_dispersion}
\begin{figure}[!t]
  \centering
  \includegraphics[width=\textwidth]{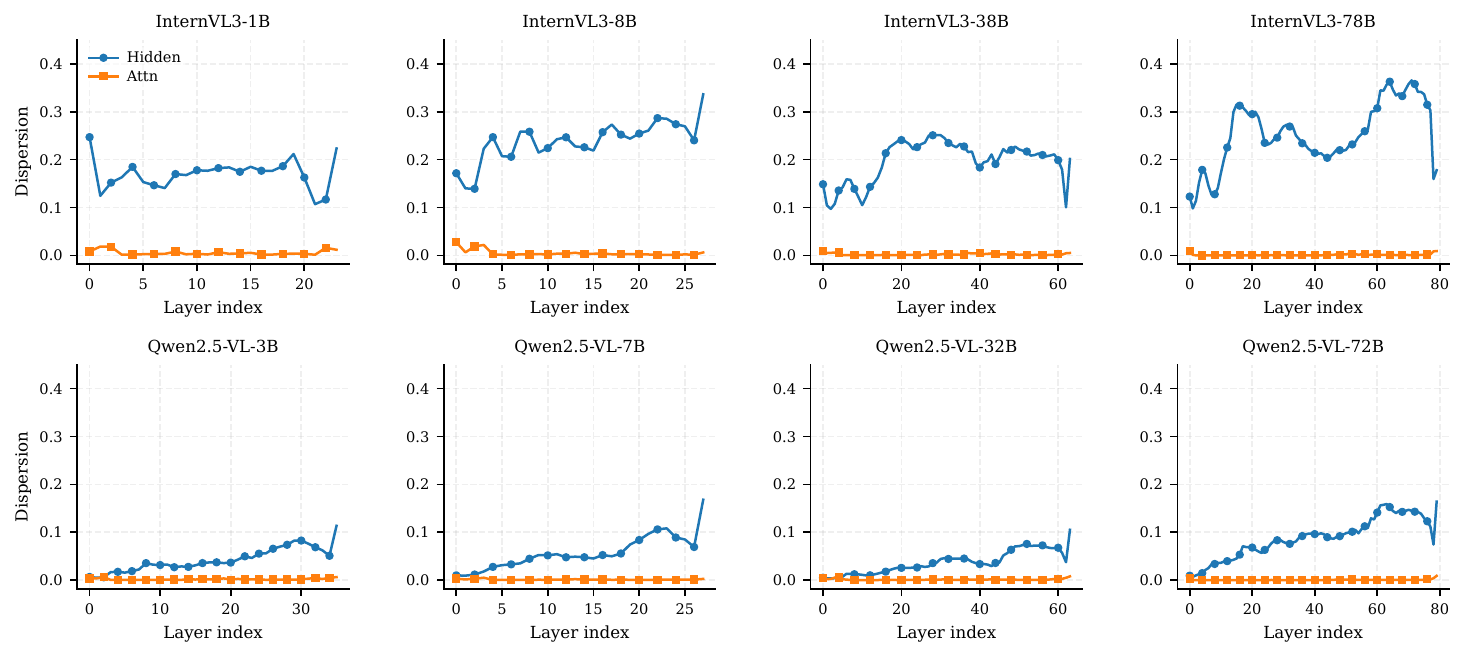}
  \caption{Hidden-state vs. attention dispersion under rewording for InternVL3 and Qwen2.5-VL model families. Each subplot corresponds to one model; the x-axis is the Transformer layer index and the y-axis is dispersion measured as $1-\cos(\cdot,\cdot)$ on the all-capabilities aggregate. We report dispersion curves for the last-token hidden representation (Hidden) and the head-mean attention map (Attn). Lower values indicate more stable signals across different question rephrasings on the same image.}
  \label{fig:appendix-hidden-vs-attn-all}
\end{figure}

We provide dispersion comparisons between hidden states and attention maps under question rewording (\cref{fig:appendix-hidden-vs-attn-all}) to justify using attention as a more capability-stable distillation signal. 

\subsection{Position-wise output-distribution alignment analyses}
\label{app:prestudy_kl}
\begin{figure}[!t]
  \centering
  \includegraphics[width=\textwidth]{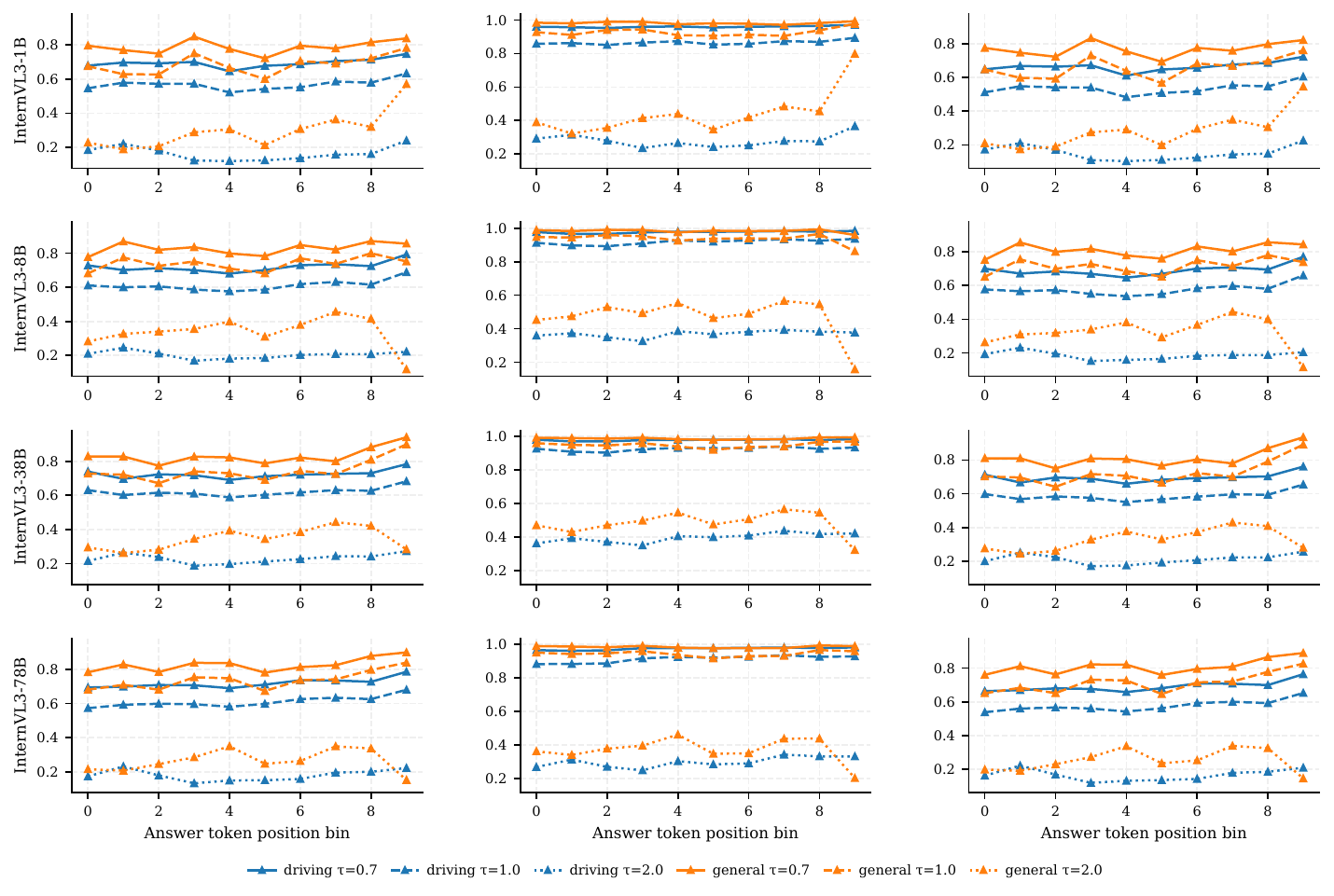}
  \caption{
    Output-distribution alignment (KL) analysis on answer-token position bins for the InternVL3 family.
    Rows correspond to models (1B/8B/38B/78B), and columns report three confidence metrics:
    top-1 probability $m$, top-10 mass $S$, and generalized margin.
    For each subplot, we compare \textit{driving} vs.\ \textit{general} answers under three temperatures
    ($\tau\in\{0.7,1.0,2.0\}$), using consistent colors (split) and line styles (temperature).
  }
  \label{fig:appendix_kl_intern}
\end{figure}

\begin{figure}[!t]
  \centering
  \includegraphics[width=\textwidth]{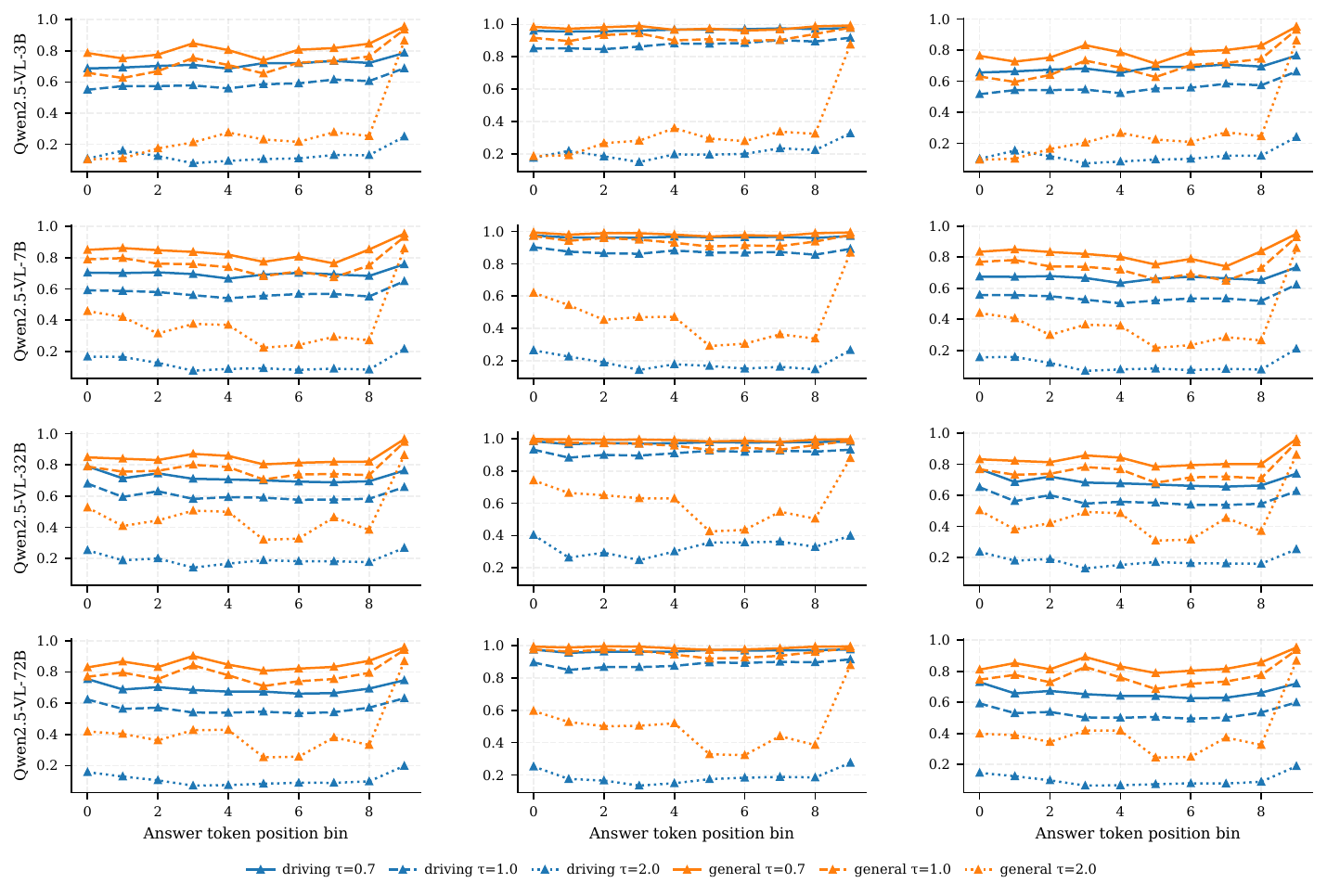}
  \caption{
    KL-based logit distillation analyses on answer-token position bins for the Qwen2.5-VL family.
    Rows correspond to models (3B/7B/32B/72B), and columns report three confidence metrics:
    top-1 probability $m$, top-10 mass $S$, and generalized margin.
    Each subplot contrasts \textit{driving} vs.\ \textit{general} answers under three temperatures
    ($\tau\in\{0.7,1.0,2.0\}$), with consistent colors (split) and line styles (temperature).
  }
  \label{fig:appendix_kl_qwen}
\end{figure}

We present position-wise analyses for output-distribution alignment under multiple temperatures (\cref{fig:appendix_kl_intern,fig:appendix_kl_qwen}), contrasting driving and general answers to clarify why KL-based output alignment can become noisy on driving outputs. 

\section{Optimization Details}
\label{app:optimization}
This section details the online loss reweighting procedure used in training, including how per-loss gradient magnitudes are estimated and how the weights are updated to stabilize multi-objective optimization. 
\subsection{Online loss reweighting strategy.}
\label{online_loss}
We use an online loss reweighting strategy to stabilize multi-objective training, balancing the influence of different losses on parameter updates and reducing sensitivity to manual tuning. 
Specifically, each loss term $L_i$ is assigned a positive learnable weight $w_i$, and we jointly optimize the weighted sum $\sum_i w_i L_i$ together with the student model.
At each optimization step, we estimate the gradient magnitude of each loss as $G_i=\lVert\nabla_\theta L_i\rVert_2$ and update $\{w_i\}$ so that the weighted magnitudes $w_i G_i$ approach a target value.
The target combines a preset static weight ratio $p_i$ and the relative decay rate of each loss, where $s$ indexes training steps:
\begin{equation}
\hat G_i \propto p_i\left(\frac{L_i(s)/L_i(0)}{\frac{1}{N}\sum_{j\in\mathcal{I}} L_j(s)/L_j(0)}\right)^{\alpha}.
\end{equation}
We encourage the weighted gradient strengths to satisfy $w_i G_i \approx \hat G_i$.
Here $\mathcal{I}$ is the set of active loss terms and $N=|\mathcal{I}|$, while $\alpha$ controls how strongly the reweighting reacts to relative loss decay.

\subsection{Teacher mixing weights.}
\label{app:teacher_mixing}
In multi-teacher training (Section~\ref{sec:multi_obj}), for a batch from capability $c\in\{p,r,pl\}$, we use the fixed teacher-mixing matrix
\begin{equation}
\Pi=
\begin{bmatrix}
0.8 & 0.1 & 0.1\\
0.1 & 0.8 & 0.1\\
0.1 & 0.1 & 0.8
\end{bmatrix}.
\end{equation}
Each row corresponds to the batch capability (perception/reasoning/planning), and each column corresponds to the perception, reasoning, and planning teachers, where each teacher defines a fixed capability-specific loss $L_t$ independent of the batch capability $c$. 
For a batch from capability $c$, the $c$-th row of $\Pi$ provides $\pi_{c,t}$ to combine the three teacher losses. 
This assigns the largest weight to the specialized teacher while keeping non-zero contributions from the other teachers.
This reduces overfitting to a single capability and mitigates catastrophic forgetting across capabilities.

\section{Annotated Distillation Data}
\subsection{Dataset Sources and Annotation Format}
Multi-view image groups are sampled from nuScenes, 
and single-view images are sampled from both nuScenes and BDD100K \cite{Caesar_2020_CVPR,bdd100k}. 
For each image (single-view) or image group (multi-view), we annotate eight driving-related question types spanning the sequential triad of perception, reasoning, and planning. 
Each instance is paired with a human reference answer consisting of a brief final response and a detailed chain-of-thought rationale. 

\subsection{Dataset Diversity Statistics}
\label{app:data_diversity}
Our human annotations span eight driving-related question types, including motion status, target identification, multi-view object extraction and attention-based reasoning, ego-behavior reasoning, rationale-grounded planning decisions, safe-action set analysis, and collision-inducing action analysis. 
The dataset spans three capability categories, and includes both single-image and six-image settings (56.23\% vs.\ 43.77\%), where the latter covers surround-view camera perspectives. 
In addition, targets are broadly distributed in the image space: center-mid 43.10\%, right-mid 13.95\%, left-mid 11.85\%, center-bottom 10.57\%, right-bottom 7.41\%, left-bottom 4.84\%, center-top 4.45\%, right-top 2.46\%, and left-top 1.37\%. 

At the object level, for road-structure and traffic-rule related objects, lane accounts for 59.60\%, traffic sign 25.90\%, and traffic light/traffic signal 9.97\%, along with long-tail elements such as crosswalk, curb, sidewalk, and stop sign. 
For vehicles, sedan accounts for 43.35\%, SUV 26.96\%, and truck/van 21.40\%, and the annotations also include bus, pickup, and taxi. 
For vulnerable road users, the annotations cover pedestrians, motorcycles, bicycles, and cyclists. 
These objects are accompanied by precise color and shape descriptions. 

At the scene level, the annotations span diverse road structures and interactions, dominated by turning (50.20\%) and intersections (14.21\%), followed by lane change (8.74\%), parking (8.55\%), and crosswalk areas (7.19\%), with additional coverage of construction, traffic jams and ramp merges. 
Weather and time-of-day mentions are also diverse (clear/rain/overcast, plus snow/wind/fog; and night/daytime/dawn--dusk). 
The action space includes both routine safe decisions and counterfactual risky behaviors: stop (26.69\%), keep speed (15.98\%), turn right/left (14.63\%/14.15\%), gentle braking to stop (7.75\%), gradual deceleration without braking (4.83\%), slight offset left/right (6.53\% total), lane change left/right (3.34\% total), and collision-oriented counterfactuals such as accelerating straight (3.40\%) as well as sustained deviation and sharp turns. 

Finally, at the language level, the average annotation length is about 61 words / 410 characters, with a clear spread (p25=48 words, p75=71 words, max=205 words), demonstrating rich variation in expression and explanatory details. 

\subsection{Annotated Examples}
\label{app:annotations}
\begin{figure}[!htbp]
  \centering
  \setlength{\fboxsep}{7pt}
  \setlength{\fboxrule}{0.4pt}

  \fbox{%
  \begin{minipage}{\dimexpr\linewidth-2\fboxsep-4\fboxrule-3pt\relax}
    \small

    \textbf{Annotated Example (Perception)}\par
    \medskip\hrule\medskip

    \centering
    \begin{minipage}{0.4\linewidth}\centering
      \includegraphics[width=\linewidth]{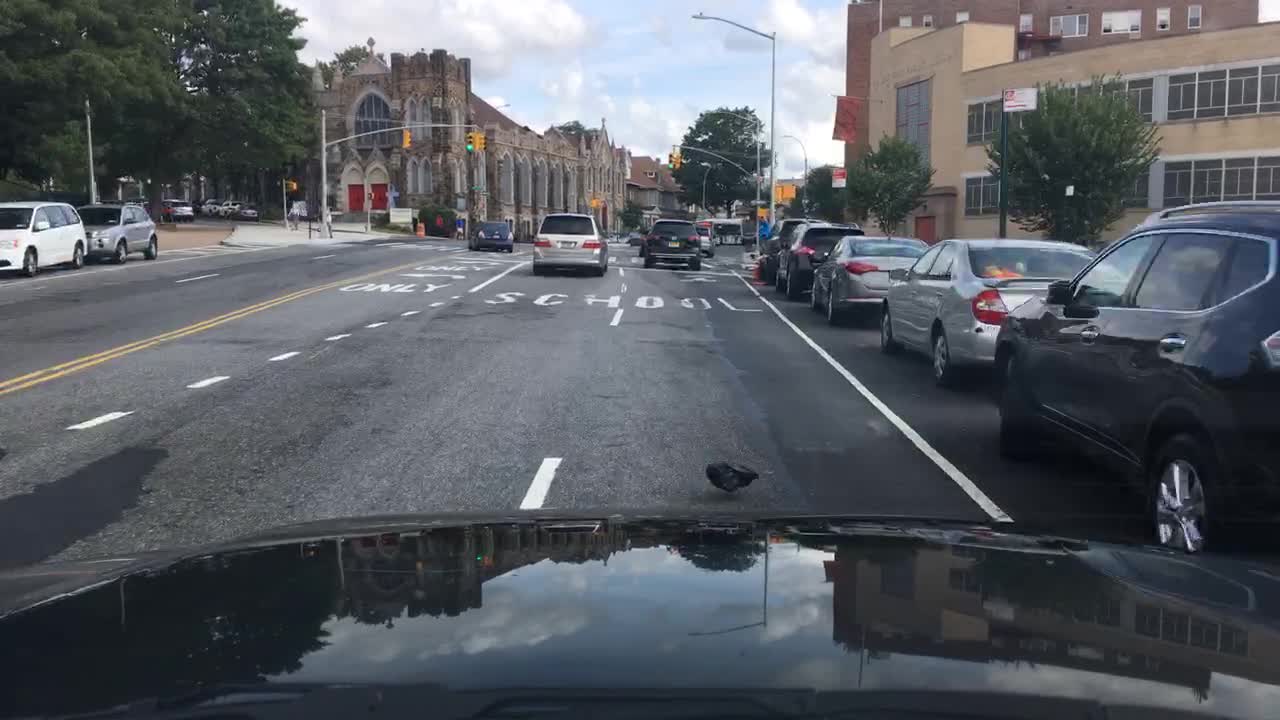}\\[-1mm]
      {\scriptsize \textbf{Front (single-view)}} 
    \end{minipage}

    \vspace{0.8em}
    \hrule
    \vspace{0.8em}

    {\ttfamily\raggedright\setlength{\parindent}{0pt}\setlength{\parskip}{3pt}%
    \begin{tabular}{@{}p{\linewidth}@{}}
\textbf{Question:} What is the moving status of object \texttt{<c1,CAM\_FRONT,0.3843,0.3338>}? Please select the correct answer from the following options:\\
A.\ Going ahead. \quad B.\ Turn left. \quad C.\ Turn right.\\[2pt]

\textbf{Answer:} B\\

\textbf{Rationale:} Objects: Black SUV (ID 1). Location \& orientation: left of ego at mid-distance; its heading shows a slight tilt to the left, indicating a left turn relative to the lane/curb. Environment: cloudy weather, dry road surface, green traffic signal. Reasoning: the observed trajectory and slight leftward tilt, together with its position in the left lane, indicate it is turning left; lane markings and alignment with the curb further support this motion state.
    \end{tabular}%
    }

    \medskip\hrule
  \end{minipage}
  }

  \caption{Perception annotation example from our 10,500 distillation dataset.}
  \label{fig:appendix_annotation_example_perception_single}
\end{figure}

\begin{figure}[!htbp]
  \centering
  \setlength{\fboxsep}{7pt}
  \setlength{\fboxrule}{0.4pt}

  \fbox{%
  \begin{minipage}{\dimexpr\textwidth-2\fboxsep-4\fboxrule-3pt\relax}
    \small

    \textbf{Annotated Example (Reasoning)}\par
    \medskip\hrule\medskip

    \centering
    \begin{tabular}{ccc}
      \begin{minipage}{0.26\textwidth}\centering
        \includegraphics[width=\linewidth]{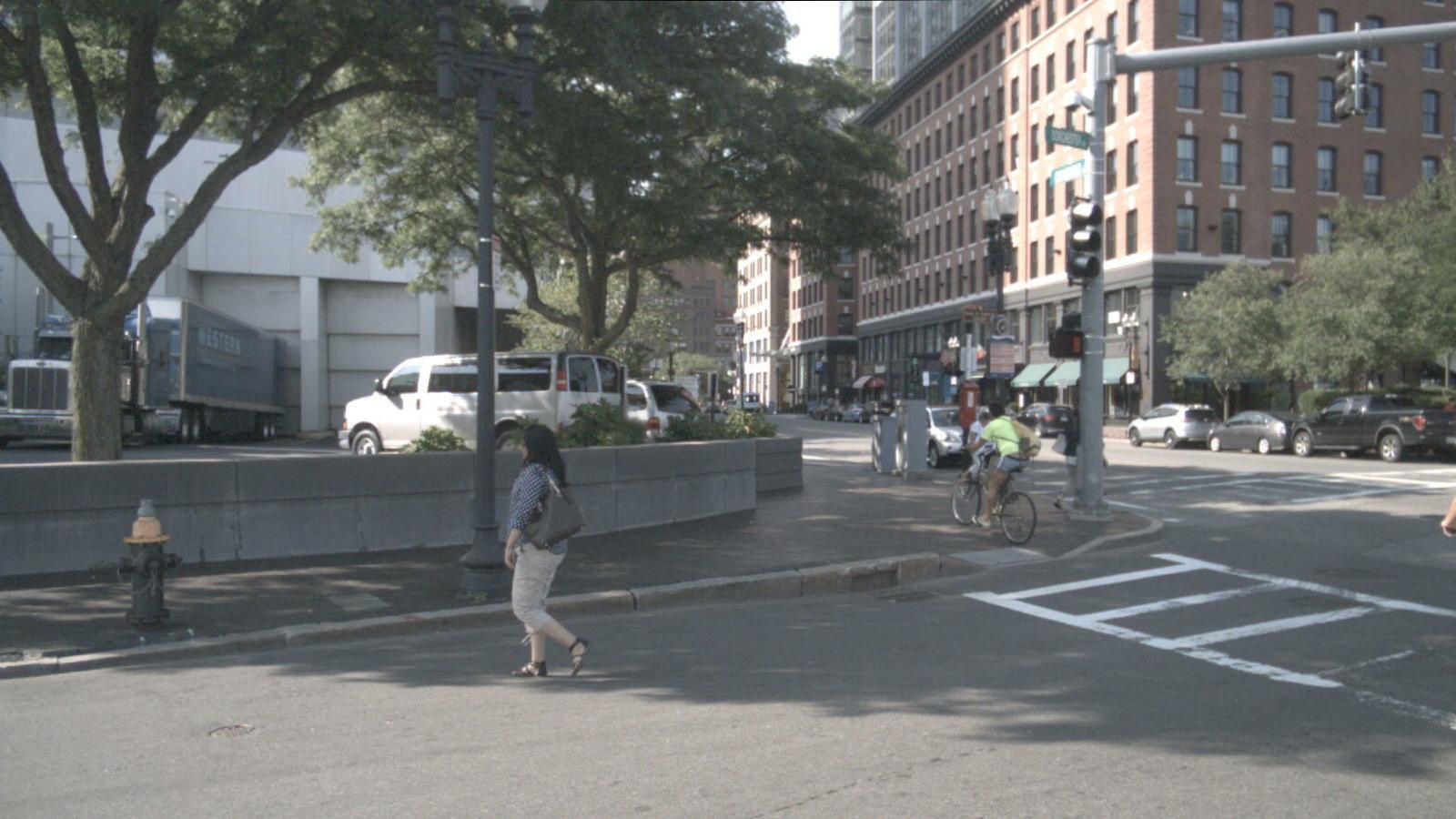}\\[-1mm]
        {\scriptsize \textbf{Front-Left}} 
      \end{minipage} &
      \begin{minipage}{0.26\textwidth}\centering
        \includegraphics[width=\linewidth]{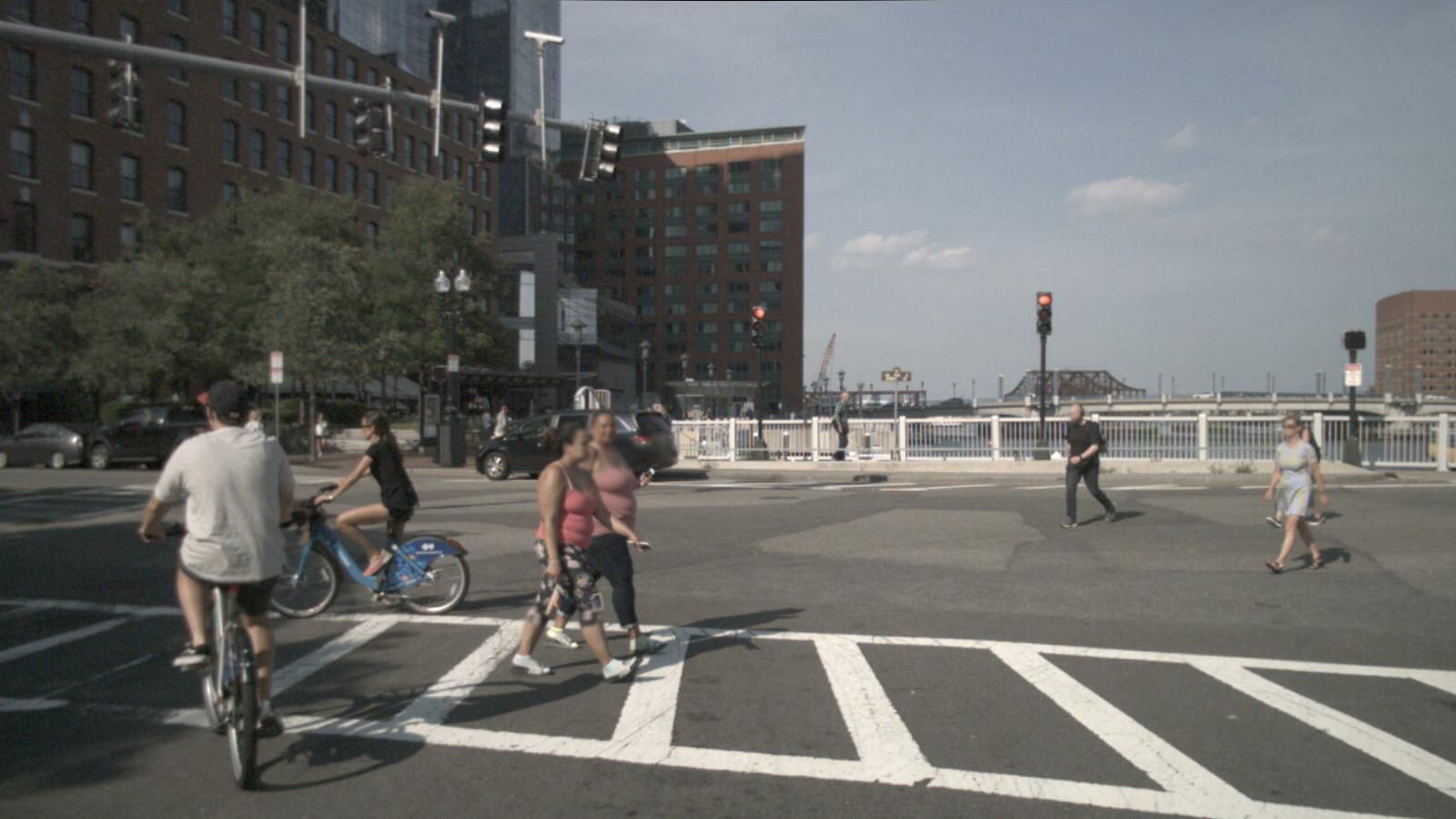}\\[-1mm]
        {\scriptsize \textbf{Front}} 
      \end{minipage} &
      \begin{minipage}{0.26\textwidth}\centering
        \includegraphics[width=\linewidth]{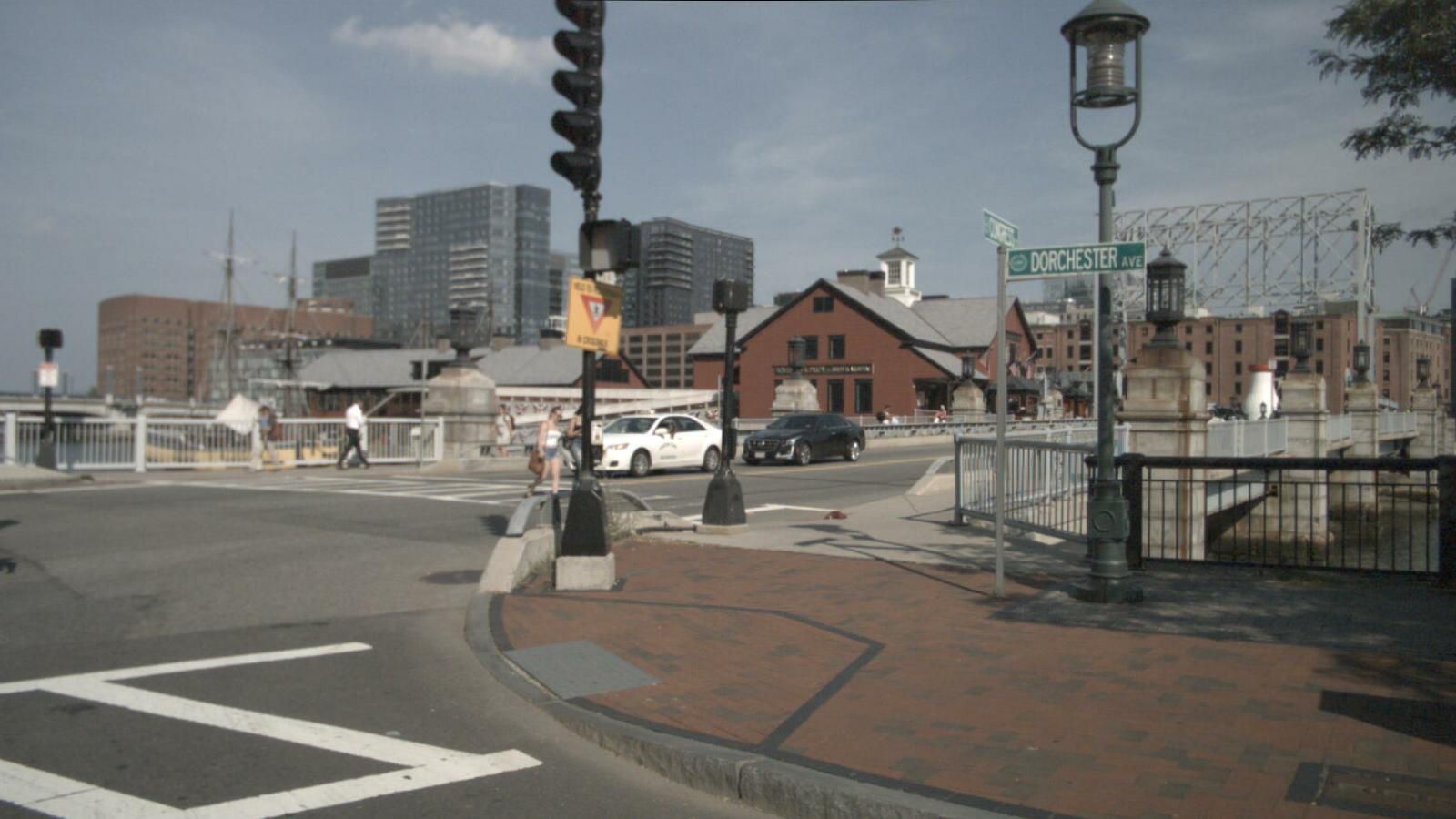}\\[-1mm]
        {\scriptsize \textbf{Front-Right}} 
      \end{minipage} \\
      \begin{minipage}{0.26\textwidth}\centering
        \includegraphics[width=\linewidth]{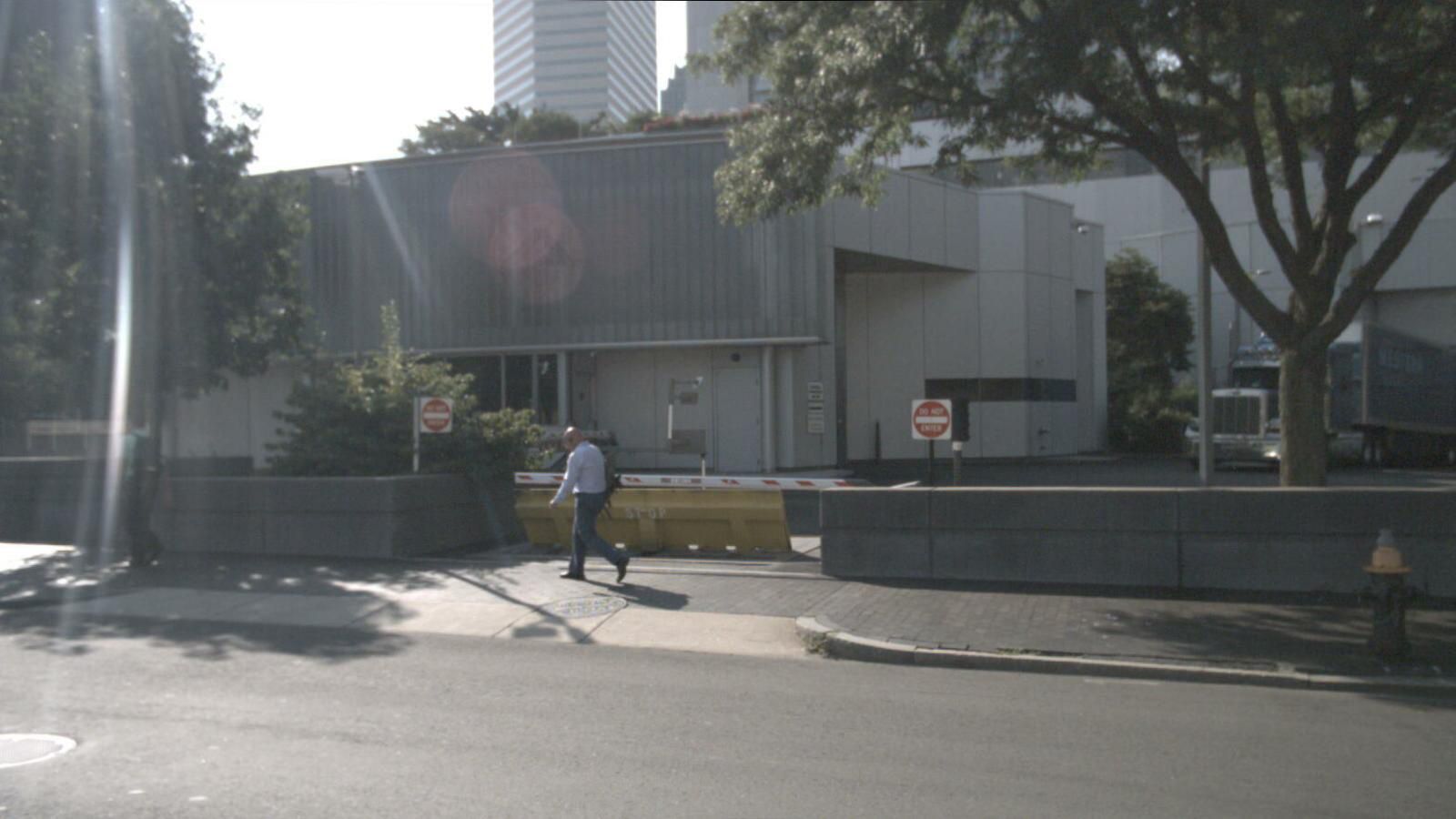}\\[-1mm]
        {\scriptsize \textbf{Back-Left}} 
      \end{minipage} &
      \begin{minipage}{0.26\textwidth}\centering
        \includegraphics[width=\linewidth]{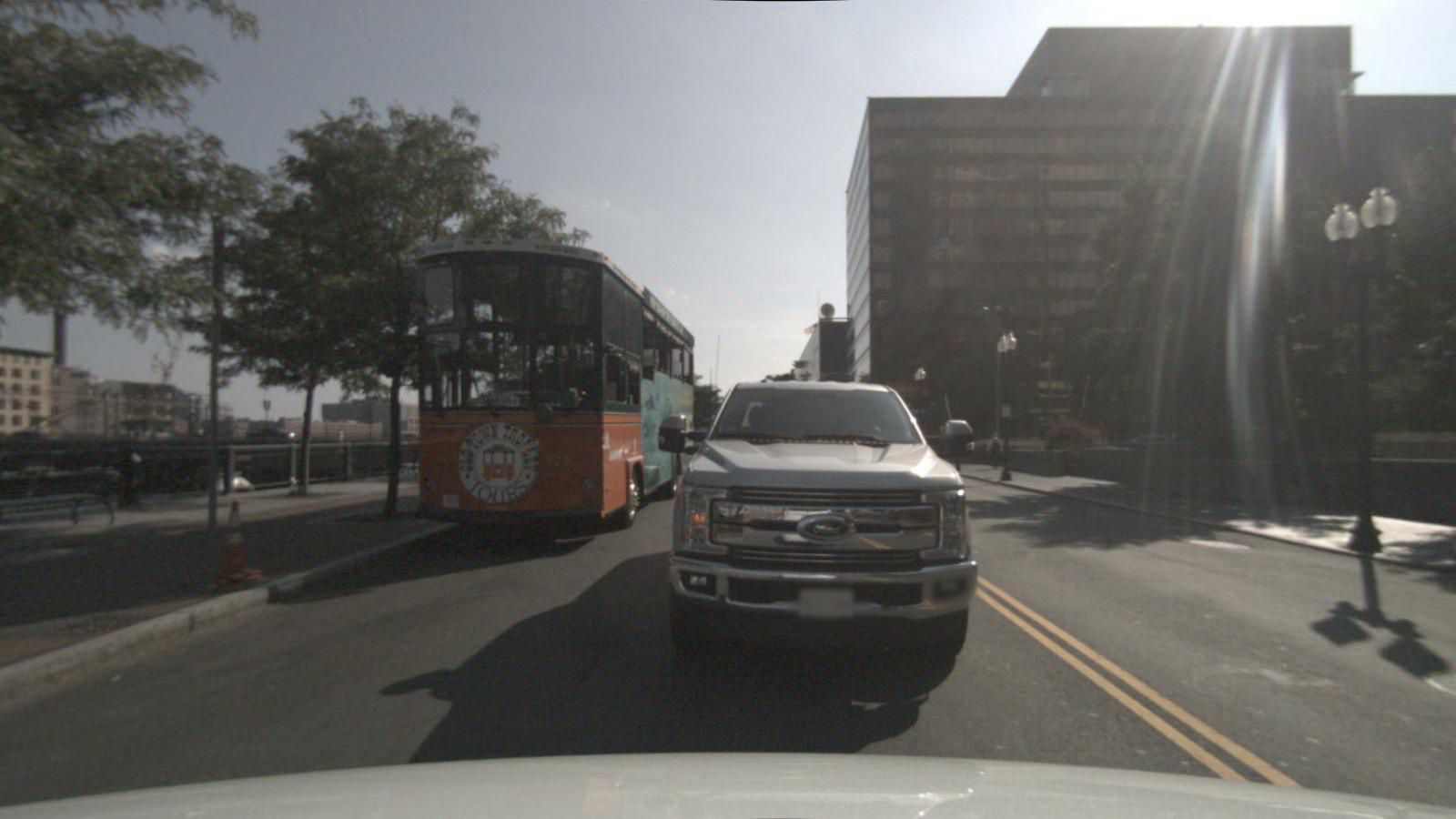}\\[-1mm]
        {\scriptsize \textbf{Back}} 
      \end{minipage} &
      \begin{minipage}{0.26\textwidth}\centering
        \includegraphics[width=\linewidth]{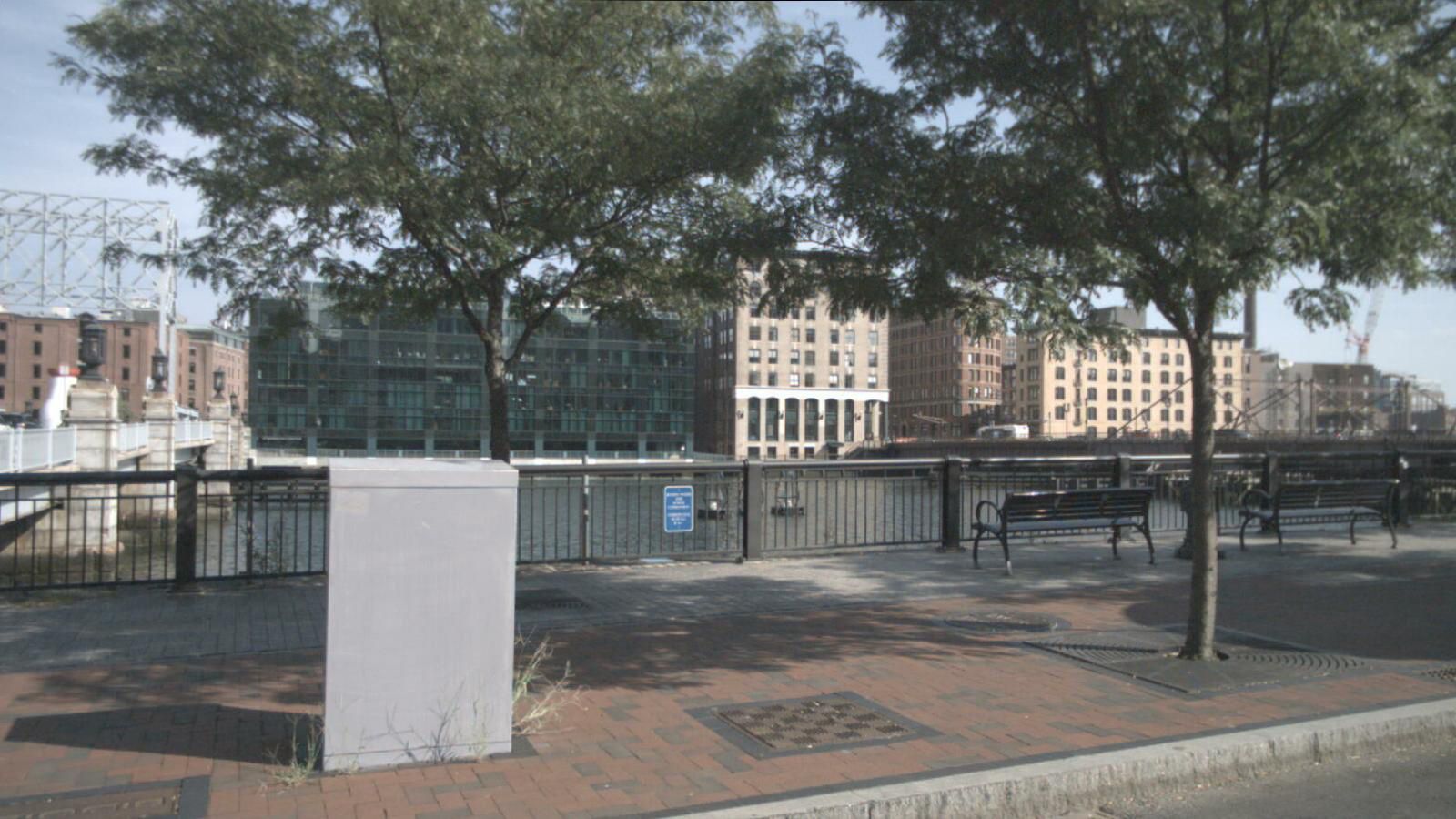}\\[-1mm]
        {\scriptsize \textbf{Back-Right}} 
      \end{minipage}
    \end{tabular}

    \vspace{0.8em}
    \hrule
    \vspace{0.8em}

    {\ttfamily\raggedright\setlength{\parindent}{0pt}\setlength{\parskip}{3pt}%
    \begin{tabular}{@{}p{\linewidth}@{}}
\textbf{Question:} Predict the behavior of the ego vehicle. Please select the correct answer from the following options: A.\ The ego vehicle is slightly steering to the right, driving with normal speed. \ B.\ \ldots \ C.\ \ldots \  D.\ \ldots \ \\[2pt]

\textbf{Answer:} A\\

\textbf{Rationale:} The vehicle is traveling at the intersection of two straight roads. Object \texttt{<own vehicle>} is located in the middle of the road, facing directly forward relative to the camera. Based on the road markings and the vehicle's orientation in the scene, its behavior is judged as \textbf{A}. The weather is clear, which does not alter the judgment.
    \end{tabular}%
    }

    \medskip\hrule
  \end{minipage}
  }

  \caption{Reasoning annotation example from our 10,500 distillation dataset.}
  \label{fig:appendix_annotation_example_reasoning}
\end{figure}

\begin{figure}[!htbp]
  \centering
  \setlength{\fboxsep}{7pt}
  \setlength{\fboxrule}{0.4pt}

  \fbox{%
  \begin{minipage}{\dimexpr\textwidth-2\fboxsep-4\fboxrule-3pt\relax}
    \small

    \textbf{Annotated Example (Planning)}\par
    \medskip\hrule\medskip

    \centering
    \begin{tabular}{ccc}
      \begin{minipage}{0.26\textwidth}\centering
        \includegraphics[width=\linewidth]{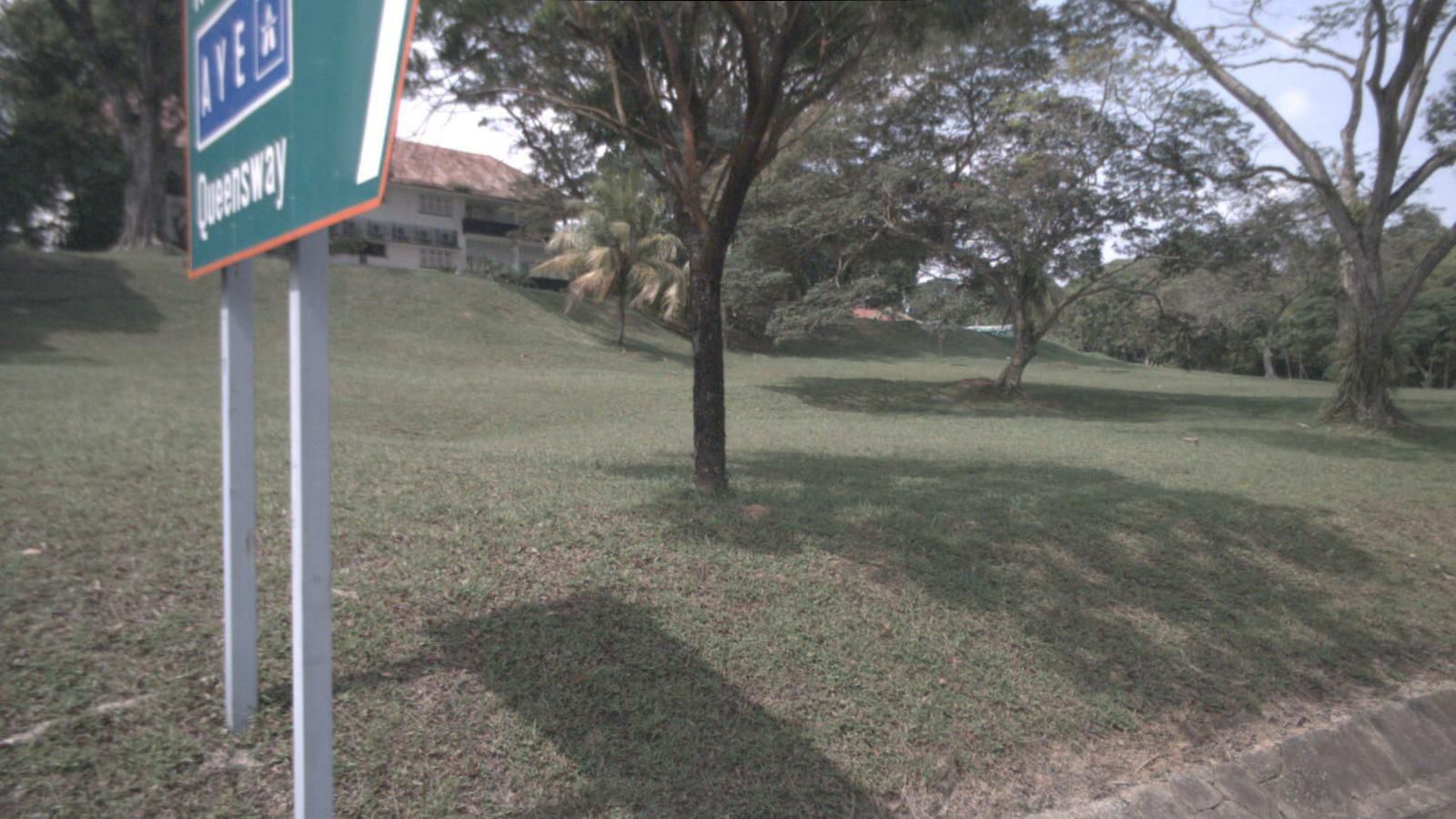}\\[-1mm]
        {\scriptsize \textbf{Front-Left}} 
      \end{minipage} &
      \begin{minipage}{0.26\textwidth}\centering
        \includegraphics[width=\linewidth]{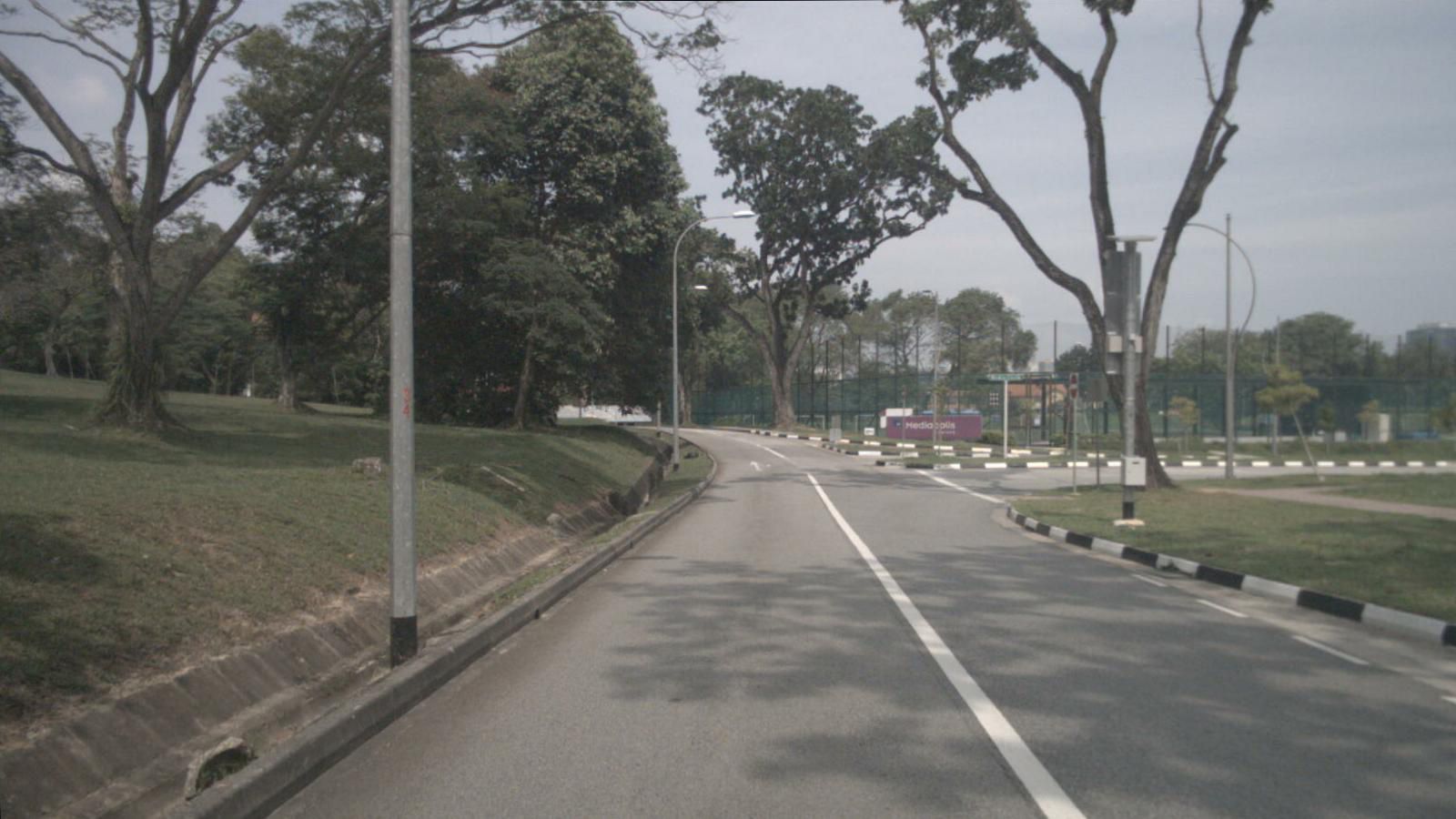}\\[-1mm]
        {\scriptsize \textbf{Front}} 
      \end{minipage} &
      \begin{minipage}{0.26\textwidth}\centering
        \includegraphics[width=\linewidth]{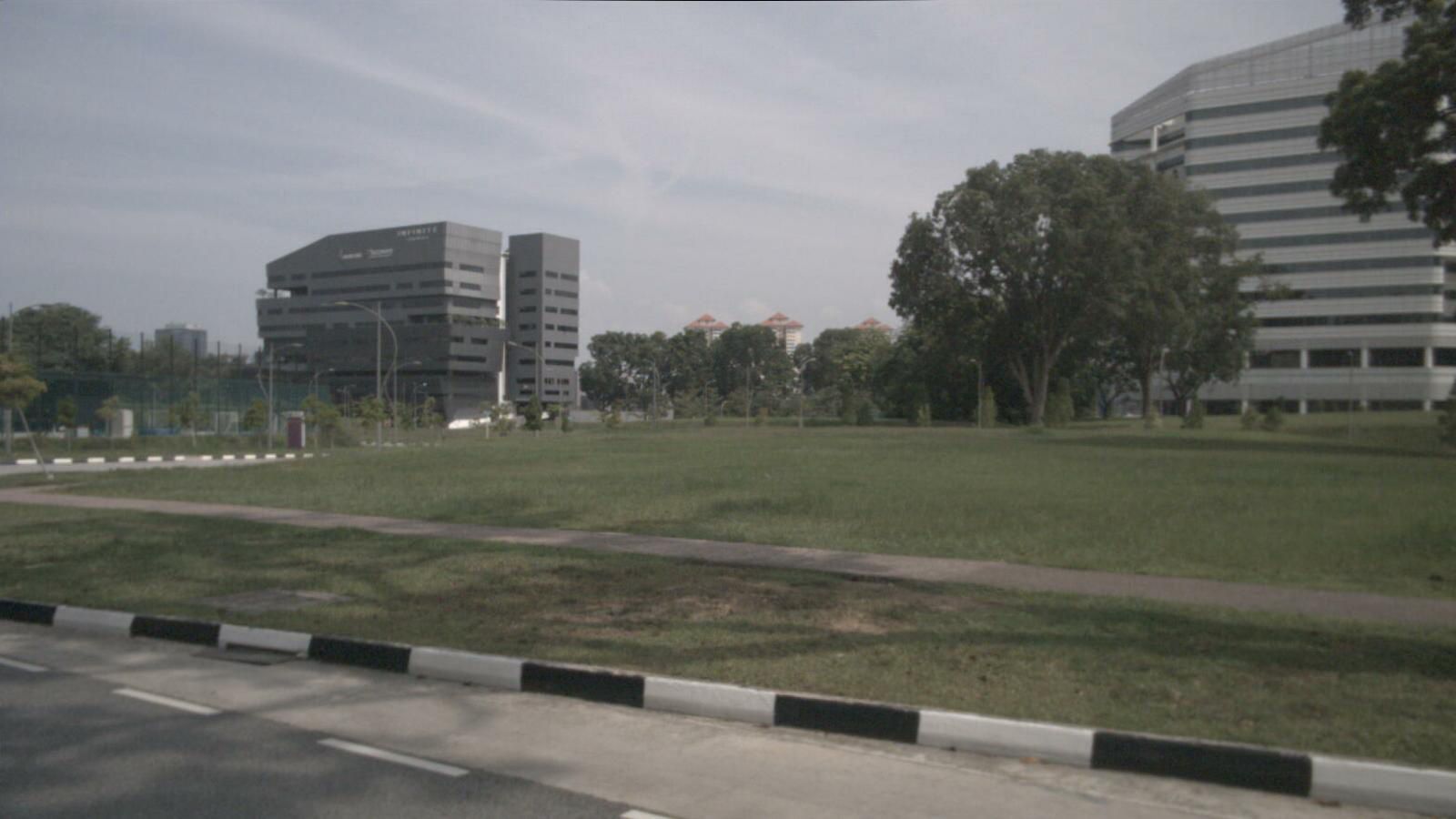}\\[-1mm]
        {\scriptsize \textbf{Front-Right}} 
      \end{minipage} \\
      \begin{minipage}{0.26\textwidth}\centering
        \includegraphics[width=\linewidth]{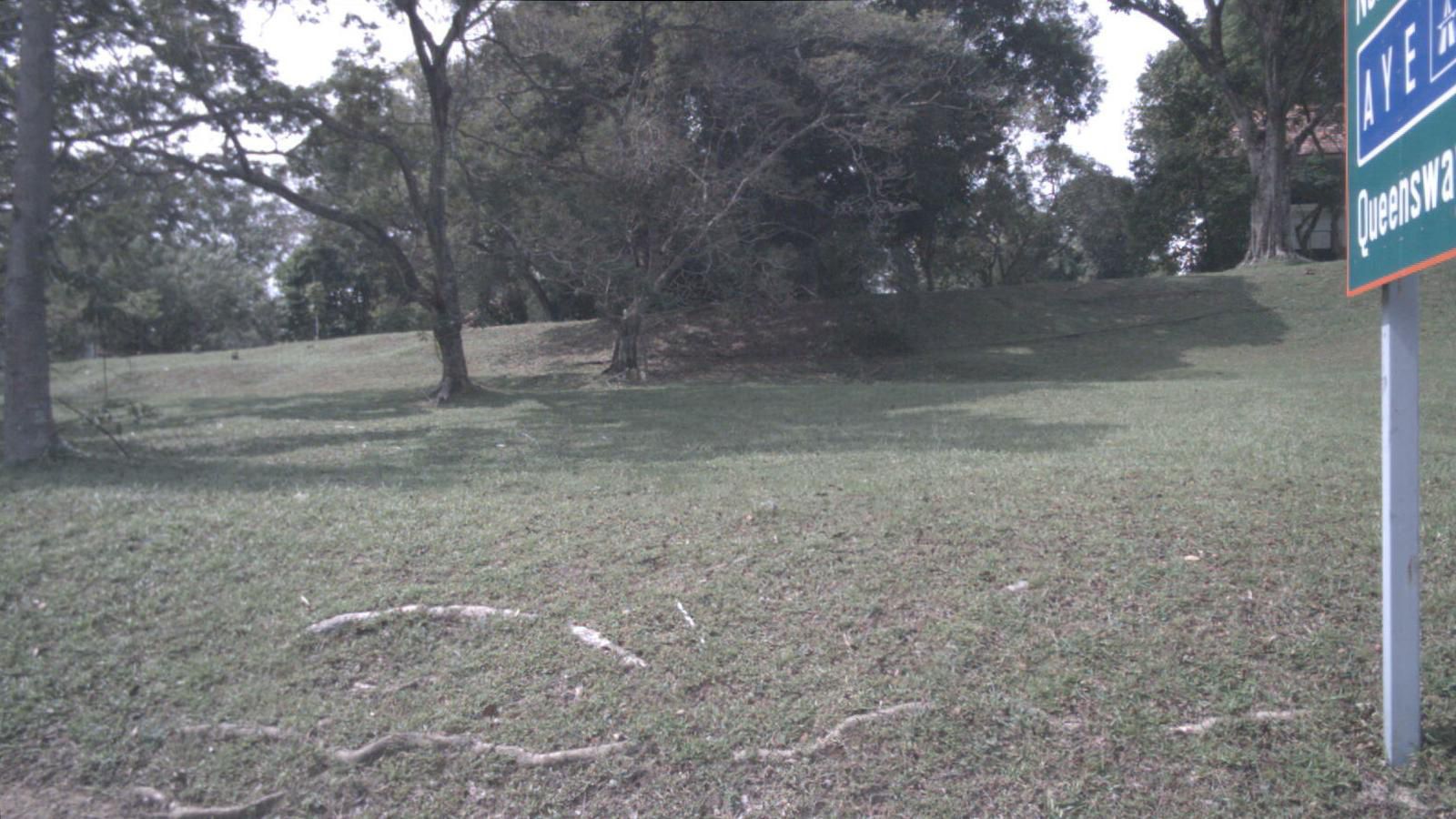}\\[-1mm]
        {\scriptsize \textbf{Back-Left}} 
      \end{minipage} &
      \begin{minipage}{0.26\textwidth}\centering
        \includegraphics[width=\linewidth]{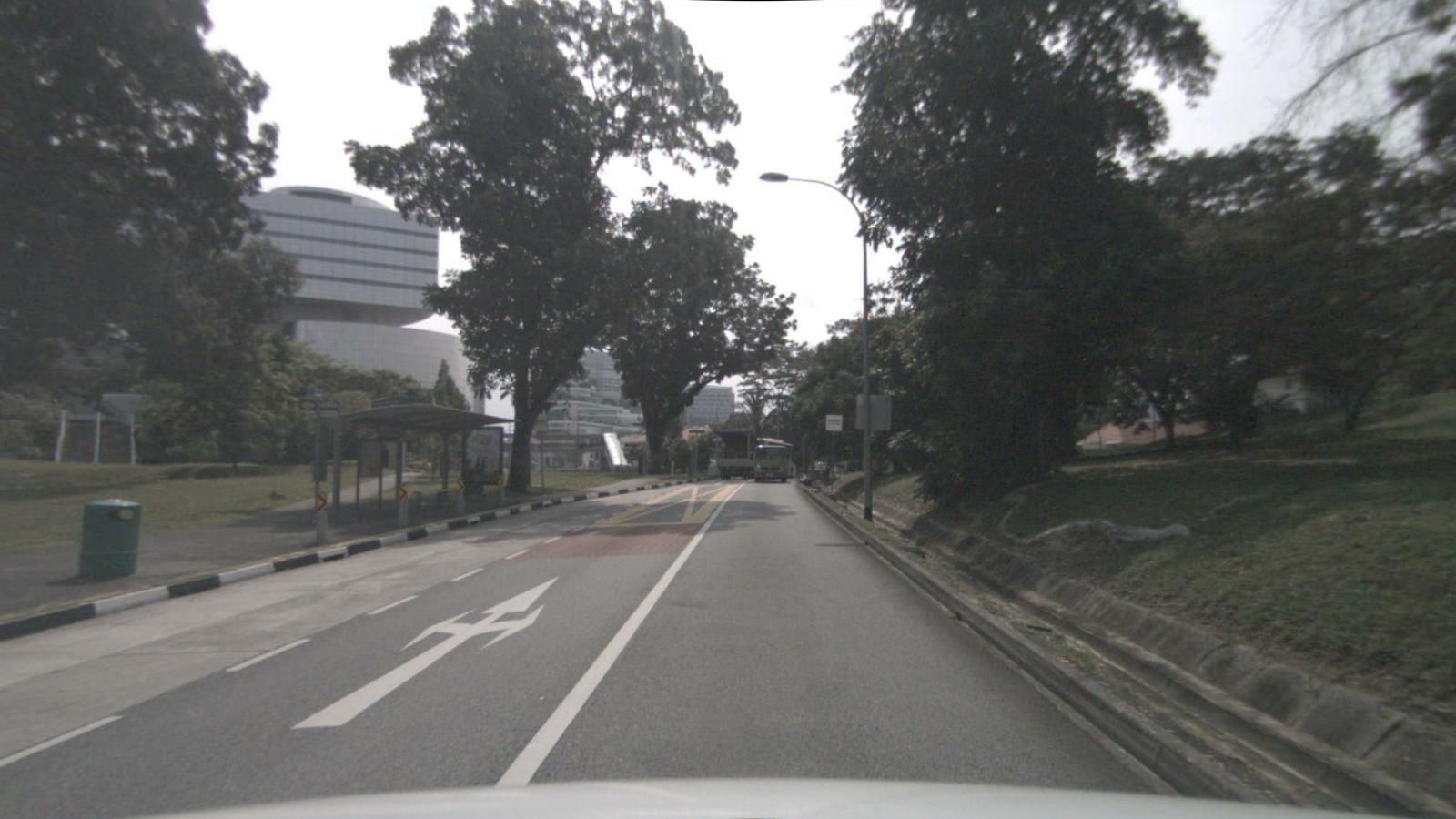}\\[-1mm]
        {\scriptsize \textbf{Back}} 
      \end{minipage} &
      \begin{minipage}{0.26\textwidth}\centering
        \includegraphics[width=\linewidth]{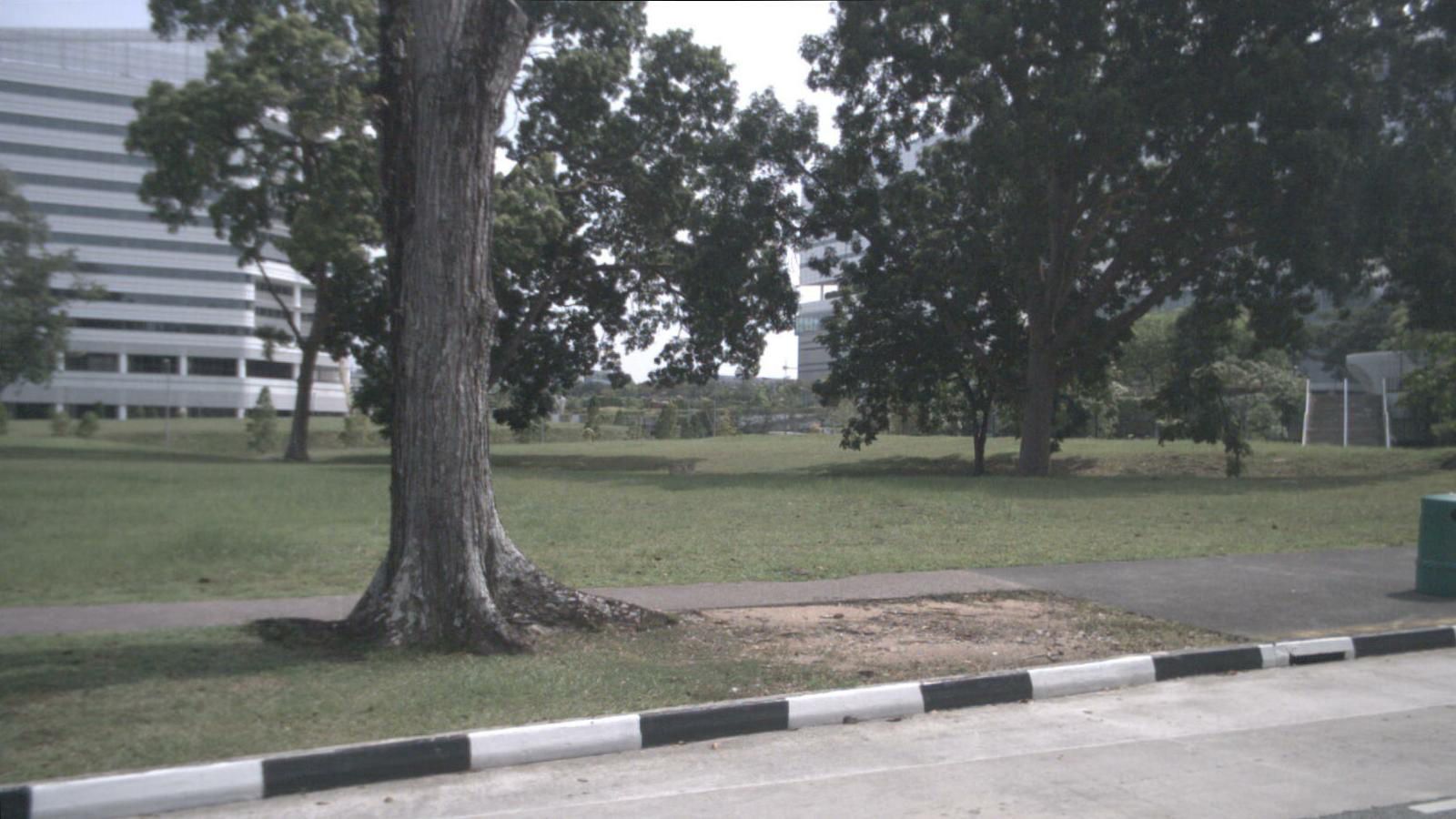}\\[-1mm]
        {\scriptsize \textbf{Back-Right}} 
      \end{minipage}
    \end{tabular}

    \vspace{0.8em}
    \hrule
    \vspace{0.8em}

    {\ttfamily\raggedright\setlength{\parindent}{0pt}\setlength{\parskip}{4pt}%
    \begin{tabular}{@{}p{\linewidth}@{}}
\textbf{Question:} In this scenario, what are safe actions to take for the ego vehicle?\\
\textbf{Answer:} keep going at the same speed; decelerate gradually without braking\\
\textbf{Rationale:} The road ahead is clear with no immediate obstacles or traffic, and the lane markings indicate a straight path. The presence of a bus in the distance suggests normal traffic flow, so maintaining speed or gently reducing it is appropriate.
    \end{tabular}%
    }

    \medskip\hrule
  \end{minipage}
  }

  \caption{Planning annotation example from our 10,500 distillation dataset.}
  \label{fig:appendix_annotation_example_planning}
\end{figure}

We group representative annotated examples for perception, reasoning, and planning (\cref{fig:appendix_annotation_example_perception_single,fig:appendix_annotation_example_reasoning,fig:appendix_annotation_example_planning}) to illustrate the format of our 10,500 distillation dataset. Each instance contains a brief answer to the question followed by a detailed chain-of-thought rationale. 

\section{Training and Evaluation Details}
\label{app:train_eval}

\subsection{Reproducibility details.}
\label{app:reproducibility}
To facilitate reproducibility, we report the full training configuration used in all main distillation experiments.

\paragraph{Training setup.}
We use full-parameter fine-tuning with a fixed random seed of 42 on NVIDIA A100-PCIE-40GB GPUs.
For the static loss coefficients, we use $w_{\mathrm{ce}}=0.7$, $w_p=60$, $w_r=0.15$, and $w_{pl}=60$ in the main experiments.
The larger perception and planning weights are used because their supervision signals are sparser and require stronger scaling for stable convergence.
We train for 3 epochs on our 10,500 human-annotated driving distillation dataset.

\paragraph{Optimization hyperparameters.}
The student model is optimized with AdamW.
We use a learning rate of $5\times 10^{-5}$.
We set gradient accumulation steps to 4.
We use a cosine learning-rate scheduler with a warmup ratio of 0.03 and a minimum learning-rate ratio of 0.01.
We set the micro-batch size to 1, and use gradient accumulation with 4 steps, resulting in an effective batch size of $1\times 4=4$.

\paragraph{Online loss reweighting implementation.}
In Appendix~\ref{online_loss}, we parameterize each learnable loss weight as $w_i=\mathrm{softplus}(a_i)+\epsilon$ with $\epsilon=10^{-8}$.
We then apply a bounded constraint to $w_i$ and use small positive epsilons for numerical stability.
The dynamic-weight parameters $\{a_i\}$ are optimized with AdamW using learning rate $3\times 10^{-4}$, with standard numerical-stability safeguards.

\paragraph{AGP numerical stability.}
In the projection operator in Section~\ref{sec:agradient_projection}, we add a small denominator constant $\epsilon_{\mathrm{proj}}=10^{-20}$ when computing $\|\mathbf{g}\|_2^2$.
This prevents division-by-zero in degenerate cases.

\subsection{Inference metrics.}
\label{app:metrics_grading}
Both training and inference are conducted on NVIDIA A100-PCIE-40GB GPUs.
Inference metrics are measured with the minimum number of GPUs required to run each model successfully.
All inference measurements are conducted on DriveBench.
We report three deployment-oriented inference metrics: peak GPU memory, average generation throughput, and median time-to-first-token (TTFT).\\
\begin{itemize}
    \item \textbf{Peak GPU memory (GB).}
We run inference with the minimum number of GPUs that can successfully execute the model, and report peak GPU memory as the maximum total memory usage summed over all GPUs observed during the run.

    \item \textbf{Average generation throughput (tokens/s).}
Average generation throughput is defined as the number of newly generated tokens divided by the total generation time:
\begin{equation}
\mathrm{avg\_gen\_tokens\_per\_s}
=
\frac{\mathrm{total\_gen\_tokens}}{\mathrm{gen\_time\_total\_s}},
\end{equation}
where $\mathrm{total\_gen\_tokens}$ excludes input prompt tokens.\\

    \item \textbf{Median TTFT (seconds).}
We measure TTFT by repeatedly running generation with $\mathrm{max\_new\_tokens}=1$ and recording the elapsed time to produce the first token. We report the median over runs:
\begin{equation}
\mathrm{ttft\_s\_median} = \operatorname{median}\left(\{T_i\}\right).
\end{equation}
\end{itemize}
\subsection{LLM-based grading protocol.}
DriveBench uses an LLM-based grader with predefined rules and scene metadata.
In our evaluation, we keep the grading prompt fixed and run grading five times, reporting the average score to reduce randomness.
We quantify the stochasticity of LLM-based grading by reporting the run-to-run standard deviation of the aggregated capability scores over the five independent grading runs.
Specifically, for each model and capability, let $\{s_i\}_{i=1}^{5}$ be the overall DriveBench score (\%) from the $i$-th grading run and $\bar s=\frac{1}{5}\sum_{i=1}^{5} s_i$ be the mean.
We report the sample standard deviation
\begin{equation}
\sigma = \sqrt{\frac{1}{5-1}\sum_{i=1}^{5}(s_i-\bar s)^2}.
\end{equation}
Table~\ref{tab:vlm_capability_efficiency_std} lists $\sigma$ for all models in Table~\ref{tab:vlm_capability_efficiency}.

\begin{table}[t]
  \caption{Run-to-run standard deviation $\sigma$ (\%) of DriveBench capability scores over five DeepSeek-V3.2 grading runs for the models in Table~\ref{tab:vlm_capability_efficiency}.}
  \label{tab:vlm_capability_efficiency_std}
  \centering
  \begin{small}
    \setlength{\tabcolsep}{6pt}
    \begin{tabular}{lccc}
      \toprule
      Model & Perception $\sigma$ & Reasoning $\sigma$ & Planning $\sigma$ \\
      \midrule
      GPT-5.1 & 1.1787 & 1.5382 & 1.4031 \\
      InternVL3-1B & 0.9494 & 1.9117 & 1.6508 \\
      InternVL3-2B & 1.3180 & 1.5707 & 1.4936 \\
      InternVL3-8B & 0.8428 & 1.6398 & 1.5451 \\
      InternVL3-14B & 0.8410 & 1.6294 & 1.6342 \\
      InternVL3-38B & 1.1267 & 1.6408 & 1.5969 \\
      InternVL3-78B & 1.2724 & 1.5285 & 1.7304 \\
      Qwen2.5-VL-3B-Instruct & 1.2111 & 1.7037 & 1.3293 \\
      Qwen2.5-VL-7B-Instruct & 1.3538 & 1.7018 & 1.2906 \\
      Qwen2.5-VL-32B-Instruct & 0.8797 & 1.9699 & 1.6058 \\
      Qwen2.5-VL-72B-Instruct & 1.2711 & 1.9081 & 1.5641 \\
      Llama-3.2-11B-Vision-Instruct & 1.3626 & 1.7238 & 1.5739 \\
      Llama-3.2-90B-Vision-Instruct & 1.2834 & 1.8497 & 1.7649 \\
      InternVL3-1B (Single) & 1.1445 & 1.8949 & 1.2617 \\
      Qwen2.5-VL-3B-Instruct (Single) & 0.9520 & 1.6770 & 1.2826 \\
      InternVL3-1B (Multi) & 0.9547 & 1.5781 & 1.4049 \\
      Qwen2.5-VL-3B-Instruct (Multi) & 1.1767 & 1.7166 & 1.4617 \\
      \bottomrule
    \end{tabular}
  \end{small}
\end{table}

\subsection{Evaluation-case illustration.}
\label{app:eval_case_illustration}
\begin{figure}[!t]
  \centering
  \setlength{\fboxsep}{7pt}
  \setlength{\fboxrule}{0.4pt}

  \fbox{%
  \begin{minipage}{\dimexpr\textwidth-2\fboxsep-4\fboxrule-3pt\relax}
    \footnotesize

    \textbf{Evaluation Example (Perception)}\par
    \medskip\hrule\medskip

    \centering
    \begin{minipage}{0.45\textwidth}\centering
      \includegraphics[width=\linewidth]{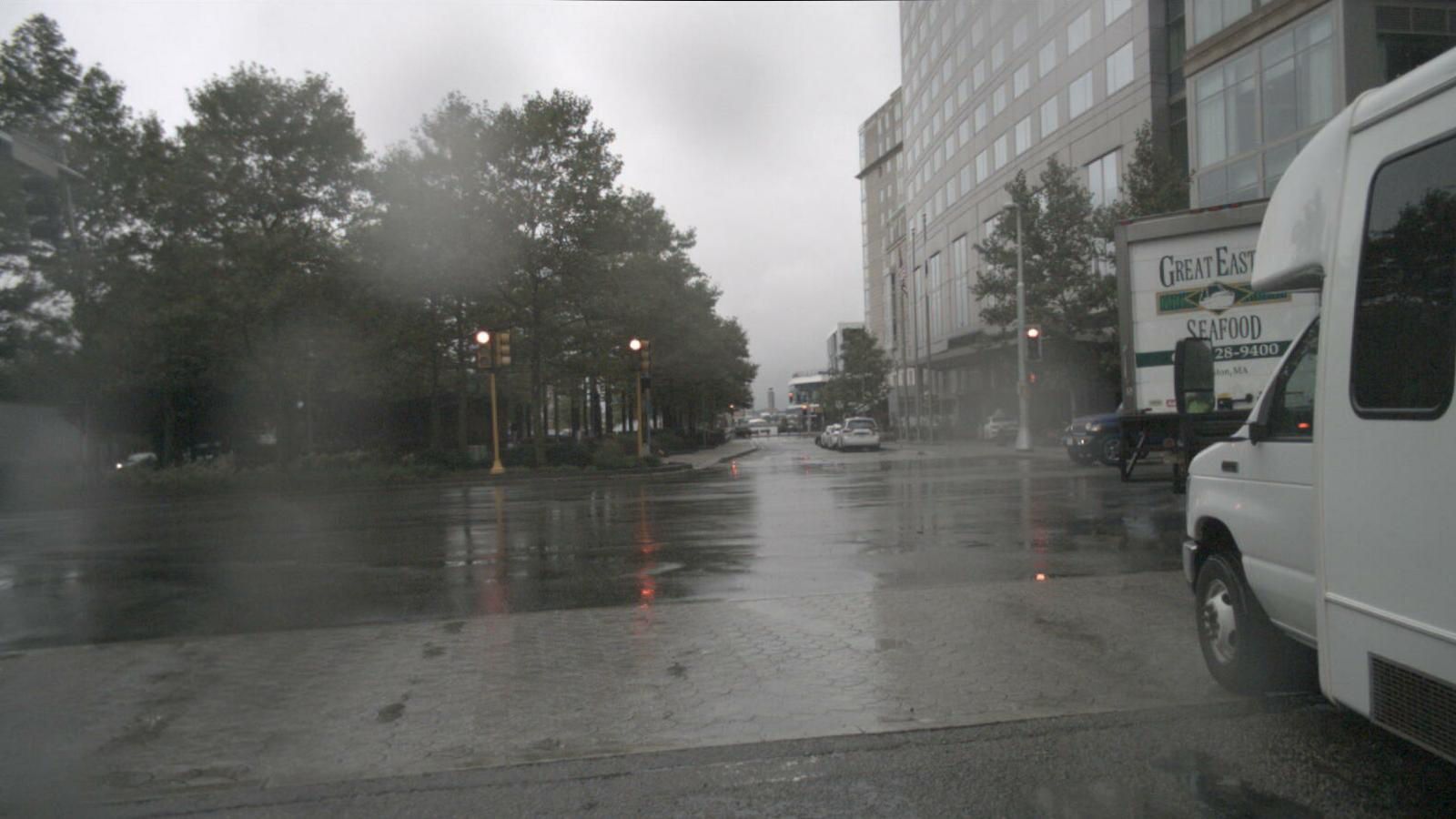}\\[-1mm]
      {\scriptsize \textbf{CAM\_FRONT}}
    \end{minipage}

    \vspace{0.8em}
    \hrule
    \vspace{0.8em}

    {\ttfamily\raggedright\setlength{\parindent}{0pt}\setlength{\parskip}{4pt}%
    \begin{tabular}{@{}p{\linewidth}@{}}
\textbf{Question:} What is the moving status of object \texttt{<c3,CAM\_FRONT,0.4755,0.5231>}? Please select the correct answer from the following options: A.\ Turn right. B.\ Turn left. C.\ Going ahead.
    \end{tabular}%
    }

    \vspace{0.6em}
    \hrule
    \vspace{0.6em}

    \begin{tabular}{@{}p{0.485\textwidth}@{\hspace{0.02\textwidth}}p{0.485\textwidth}@{}}
      {\ttfamily\raggedright\setlength{\parindent}{0pt}\setlength{\parskip}{3pt}%
      \textbf{Model:}%
      \hspace{-1pt}%
      \raisebox{-0.25\height}{\includegraphics[height=1.2em]{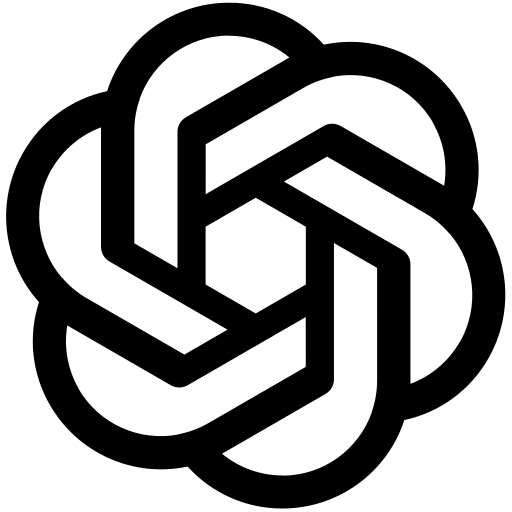}}%
      \hspace{4pt}%
      \textbf{GPT-5.1}\par
      \textbf{Prediction (compressed):}\par
      Answer: C.\ Going ahead.\par
      Explanation: Object c3 is the silver van located roughly in the center lane ahead. Its orientation aligns with the lane's straight arrow markings, and its motion appears to follow the lane forward rather than turning left or right.\ldots
      }
      &
      {\ttfamily\raggedright\setlength{\parindent}{0pt}\setlength{\parskip}{3pt}%
      \textbf{Model:}%
      \raisebox{-0.33\height}{\includegraphics[height=2.0em]{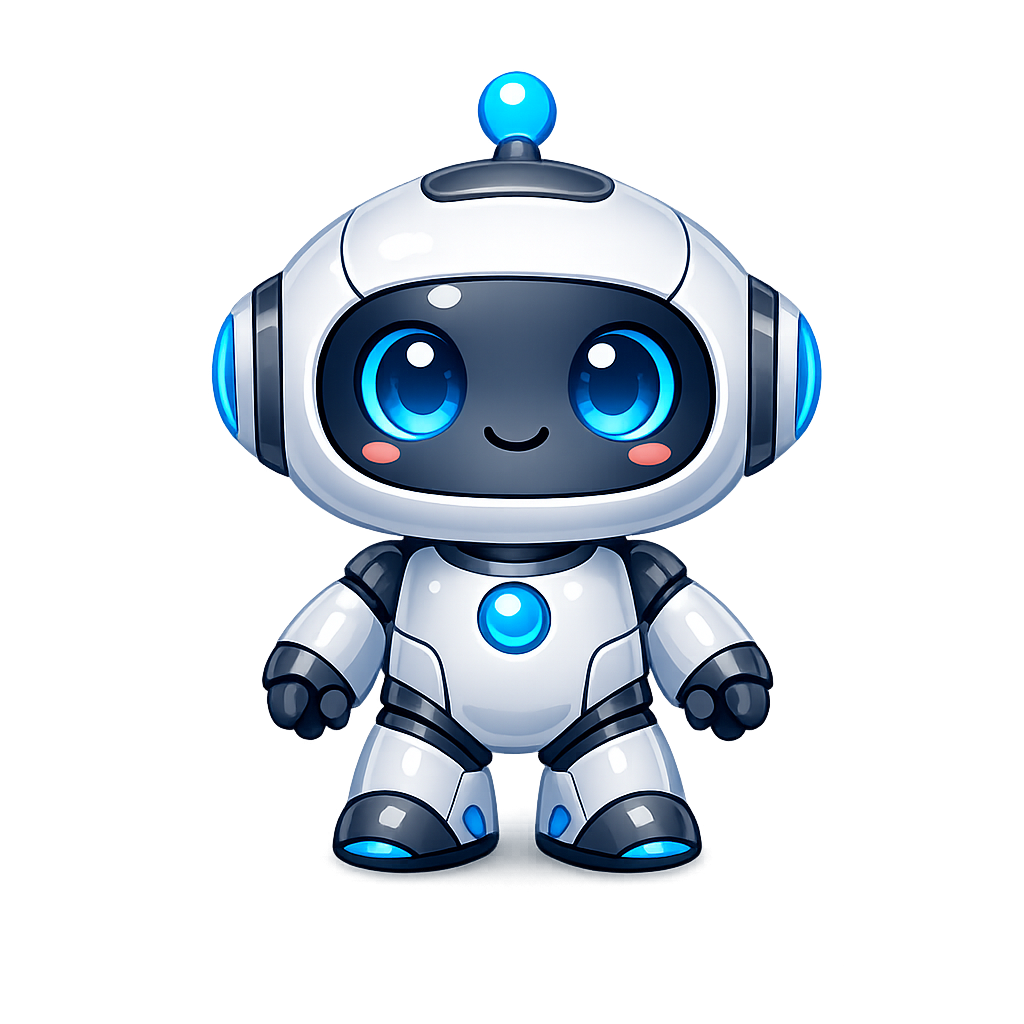}}%
      \hspace{-3.5pt}%
      \ \textbf{Ours (Drive-KD)}\par
      \textbf{Prediction:}\par
      Final: A.\ Turn right.\par
      Object: gray van\par
      Position \& orientation: relative to ego front, heading turning right\par
      Environment: visible intersection with traffic signals and lane markings\par
      Contextual evidence: the van's trajectory follows the right-turn lane, and its front wheels are angled right, \ldots
      }
    \end{tabular}

    \vspace{0.8em}
    \hrule
    \vspace{0.8em}

    {\ttfamily\raggedright\setlength{\parindent}{0pt}\setlength{\parskip}{3.5pt}%
    \begin{tabular}{@{}p{\linewidth}@{}}
\textbf{Evaluator:}%
\hspace{0.6em}%
\raisebox{-0.25\height}{\includegraphics[height=1.25em]{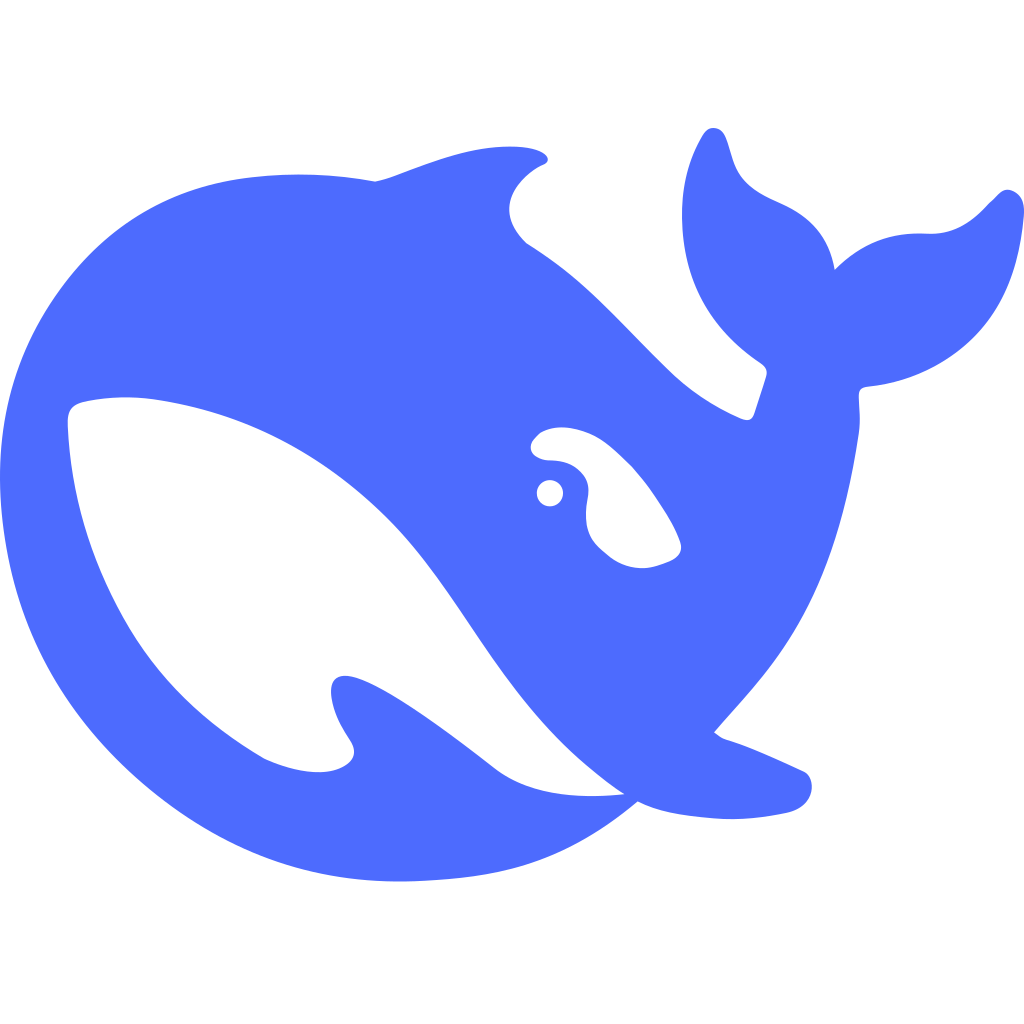}}%
\hspace{0.6em}%
\textbf{DeepSeek-V3.2}\\
\textbf{Ground Truth:} A\\[3pt]

\textbf{DeepSeek grading for GPT-5.1:}\\
1.\ Answer Correctness (50 points): 0\par
2.\ Object Recognition (10 points): 5\par
3.\ Object Location and Orientation (15 points): 5\par
4.\ Environmental Condition Awareness (15 points): 0\par
5.\ Clarity of Reasoning (10 points): 8\par
Total Score: 18\par
Brief Summary: \ldots\par

\textbf{DeepSeek grading for Ours (Drive-KD):}\\
1.\ Answer Correctness (50 points): 50\par
2.\ Object Recognition (10 points): 5\par
3.\ Object Location and Orientation (15 points): 10\par
4.\ Environmental Condition Awareness (15 points): 10\par
5.\ Clarity of Reasoning (10 points): 8\par
Total Score: 83\par
Brief Summary: The predicted answer correctly identifies the moving status as ``Turn right'' (A), matching the ground truth, earning full correctness points. Object recognition is partially accurate: the prediction describes a ``gray van,'' but the ground truth specifies a ``white sedan,'' leading to a deduction for misidentification of color and vehicle type, though both are vehicles in a similar context. Location and orientation are well described with details like ``relative to ego front, heading turning right'' and mention of lane markings and trajectory, \ldots\par
    \end{tabular}%
    }

    \medskip\hrule
  \end{minipage}
  }

  \caption{Illustration of our evaluation process on a DriveBench perception question. For readability, the model responses and DeepSeek justifications are compressed while preserving the grading structure and key reasons.}
  \label{fig:appendix_eval_process_example_perception}
\end{figure}

\begin{figure}[!t]
  \centering
  \setlength{\fboxsep}{7pt}
  \setlength{\fboxrule}{0.4pt}

  \fbox{%
  \begin{minipage}{\dimexpr\textwidth-2\fboxsep-4\fboxrule-3pt\relax}
    \footnotesize

    \textbf{Evaluation Example (Reasoning)}\par
    \medskip\hrule\medskip

    \centering
    \begin{tabular}{ccc}
      \begin{minipage}{0.26\textwidth}\centering
        \includegraphics[width=\linewidth]{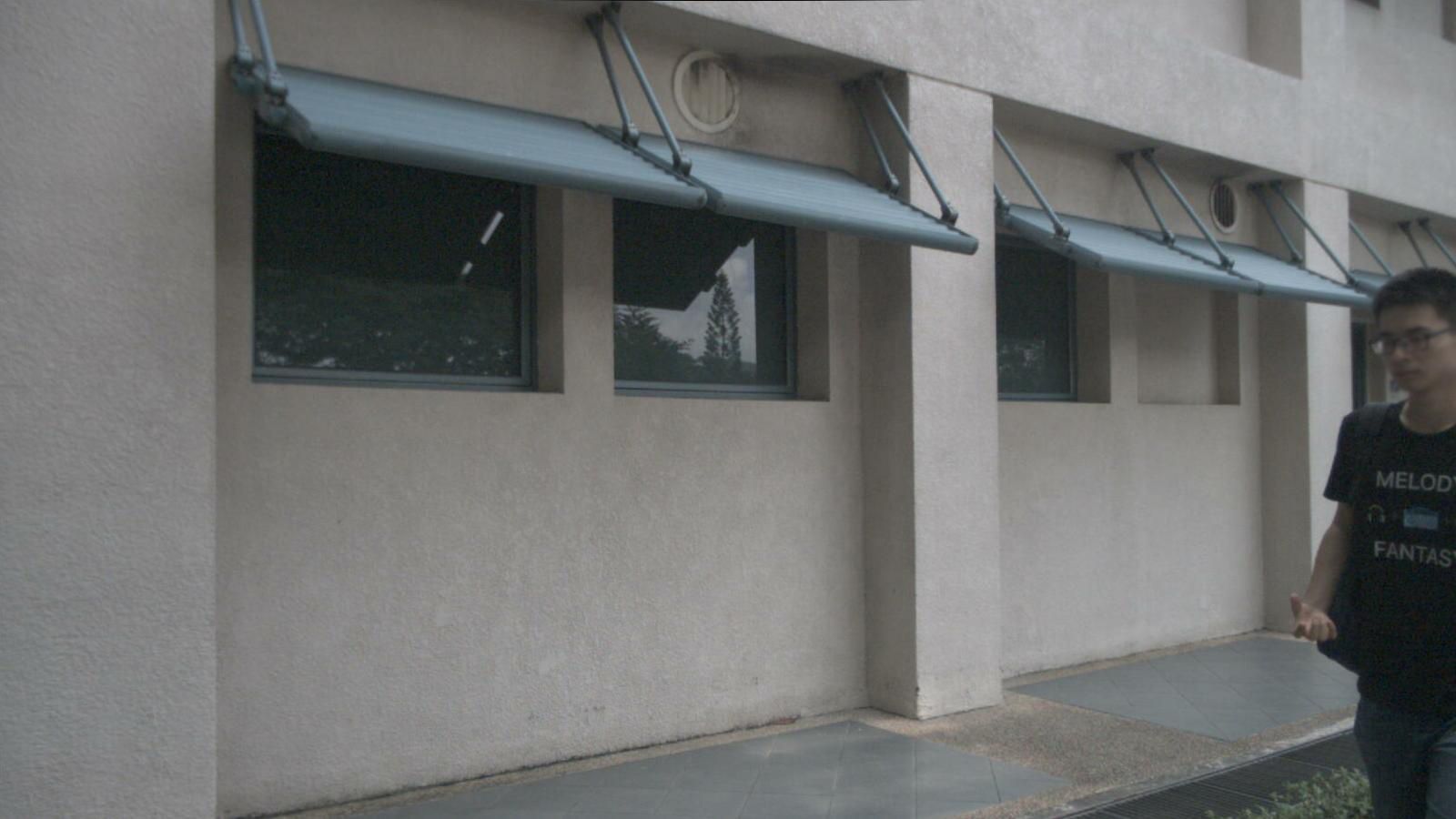}\\[-1mm]
        {\scriptsize \textbf{Front-Left}}
      \end{minipage} &
      \begin{minipage}{0.26\textwidth}\centering
        \includegraphics[width=\linewidth]{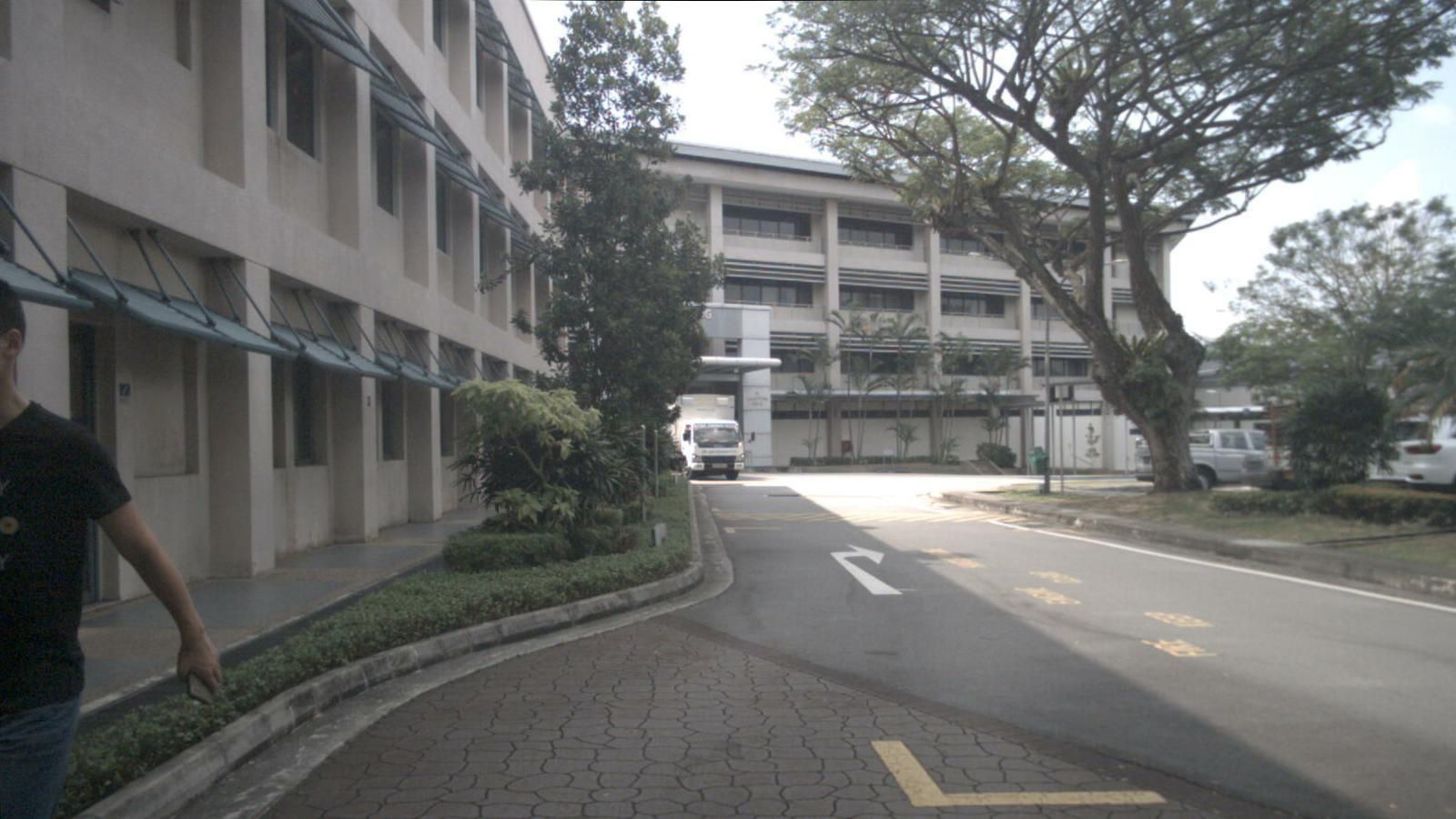}\\[-1mm]
        {\scriptsize \textbf{Front}}
      \end{minipage} &
      \begin{minipage}{0.26\textwidth}\centering
        \includegraphics[width=\linewidth]{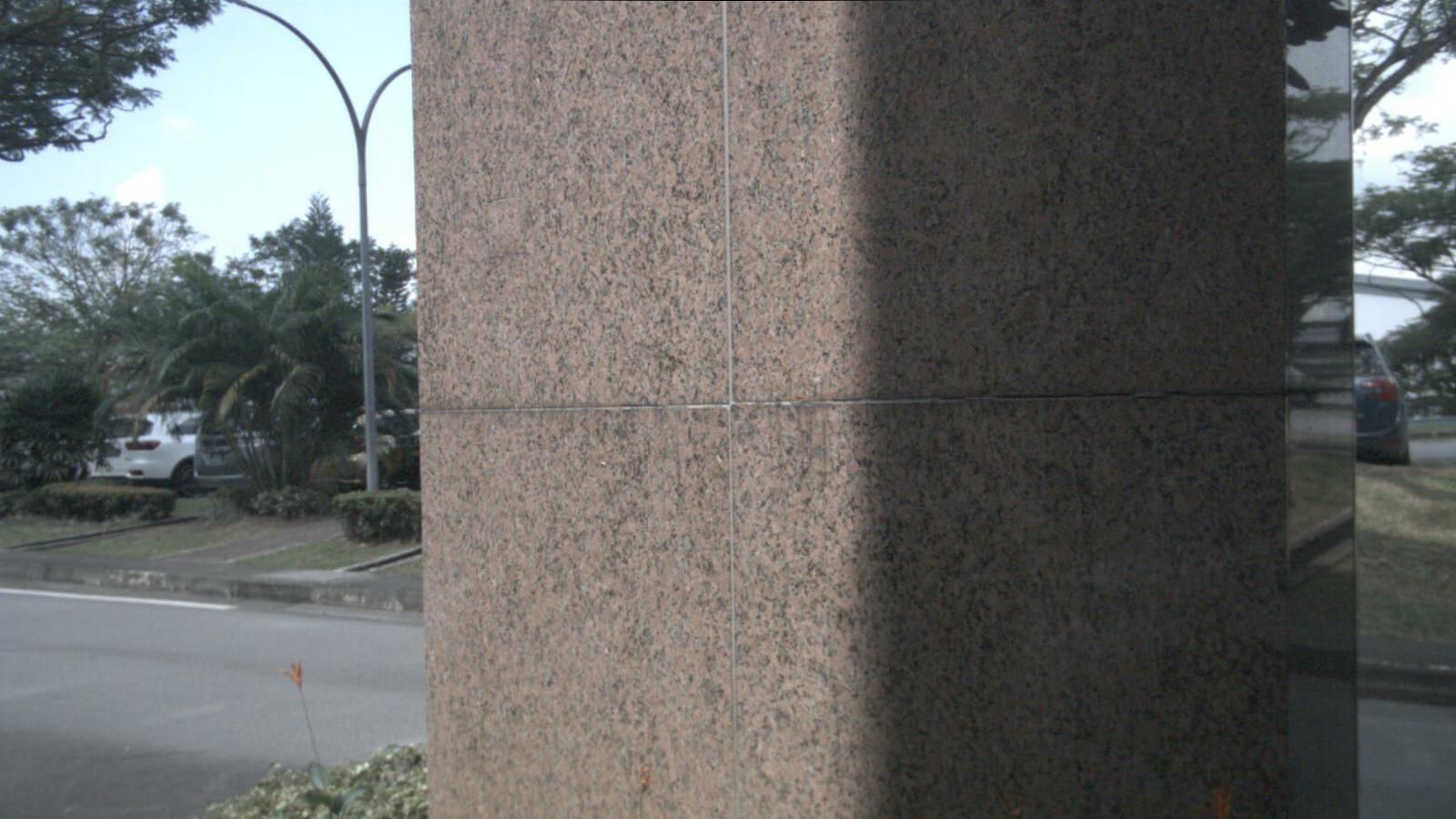}\\[-1mm]
        {\scriptsize \textbf{Front-Right}}
      \end{minipage} \\
      \begin{minipage}{0.26\textwidth}\centering
        \includegraphics[width=\linewidth]{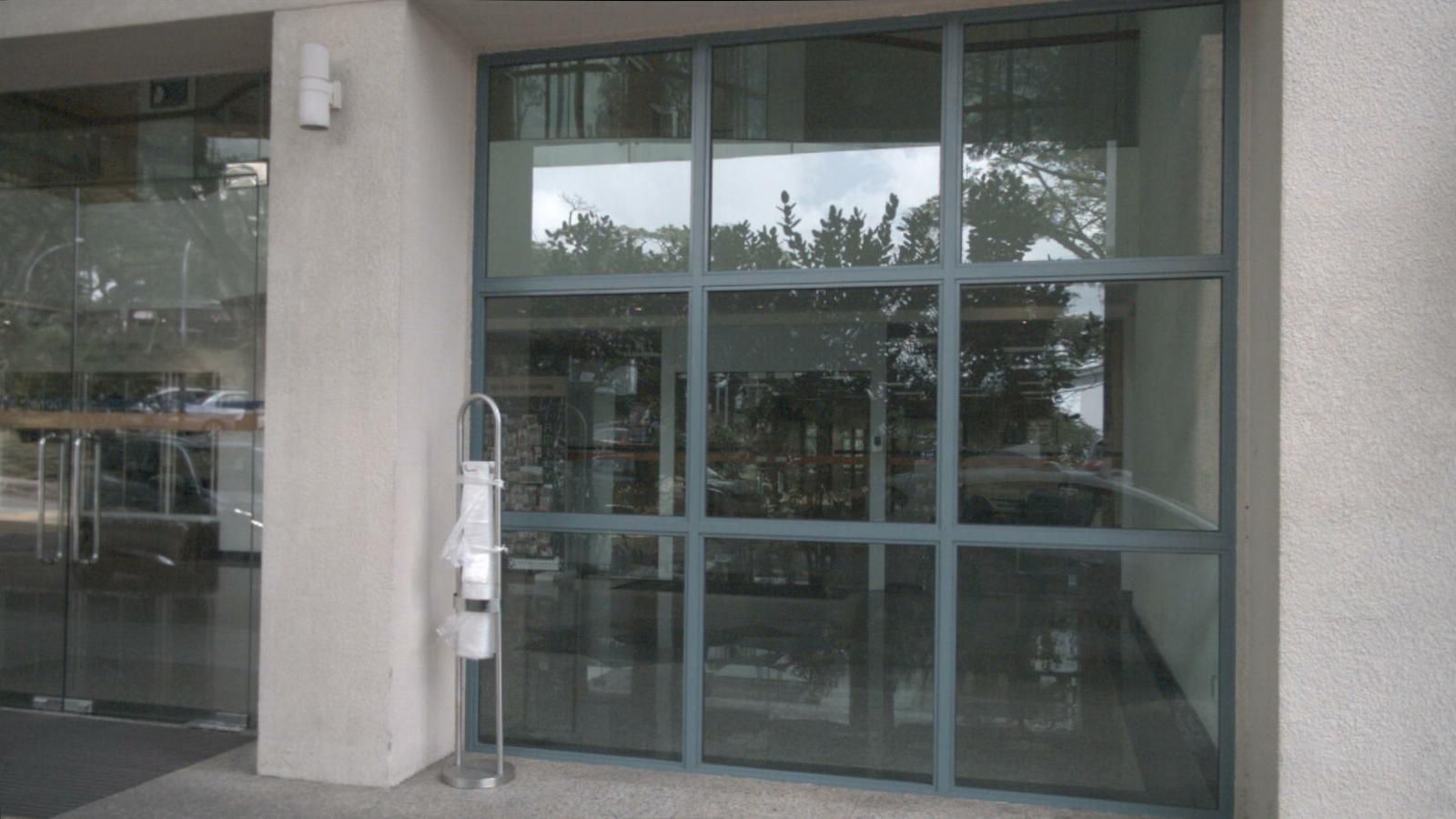}\\[-1mm]
        {\scriptsize \textbf{Back-Left}}
      \end{minipage} &
      \begin{minipage}{0.26\textwidth}\centering
        \includegraphics[width=\linewidth]{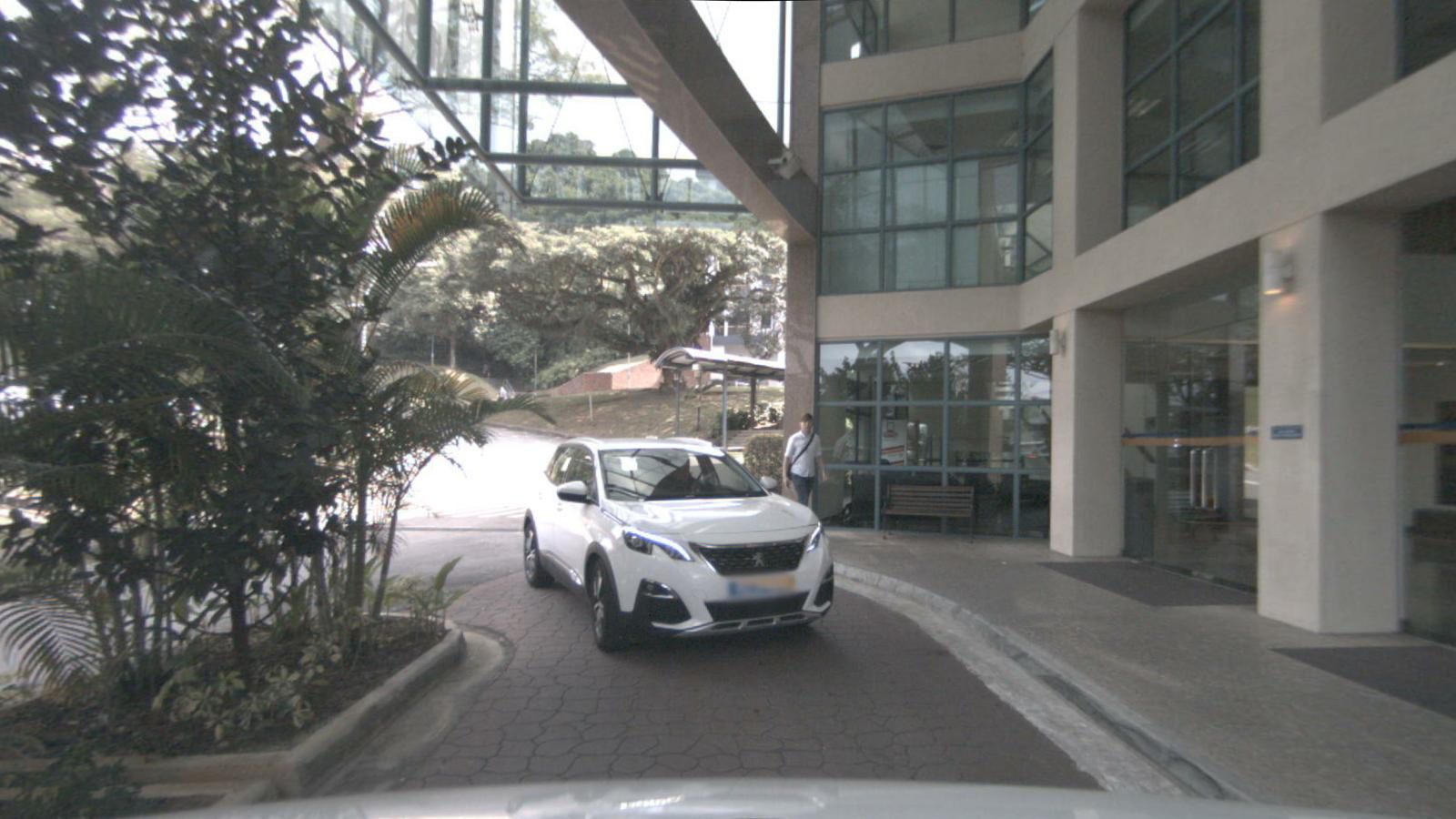}\\[-1mm]
        {\scriptsize \textbf{Back}}
      \end{minipage} &
      \begin{minipage}{0.26\textwidth}\centering
        \includegraphics[width=\linewidth]{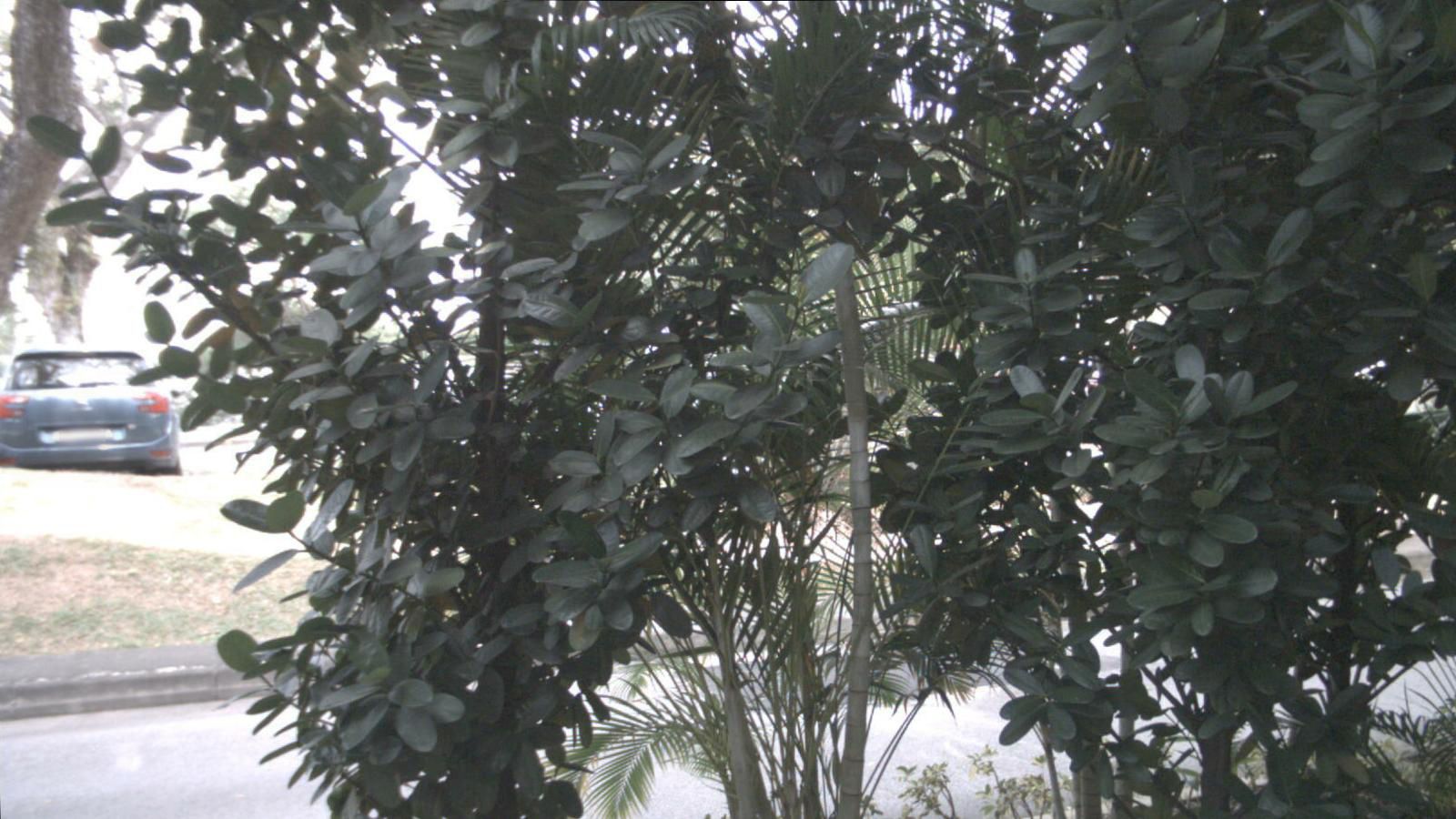}\\[-1mm]
        {\scriptsize \textbf{Back-Right}}
      \end{minipage}
    \end{tabular}

    \vspace{0.8em}
    \hrule
    \vspace{0.8em}

    {\ttfamily\raggedright\setlength{\parindent}{0pt}\setlength{\parskip}{4pt}%
    \begin{tabular}{@{}p{\linewidth}@{}}
\textbf{Question:} Predict the behavior of the ego vehicle. Please select the correct answer from the following options: A. The ego vehicle is going straight. The ego vehicle is not moving. B. The ego vehicle is steering to the right.  C. \ldots\ D. \ldots.
    \end{tabular}%
    }

    \vspace{0.6em}
    \hrule
    \vspace{0.6em}

    \begin{tabular}{@{}p{0.485\textwidth}@{\hspace{0.02\textwidth}}p{0.485\textwidth}@{}}
      {\ttfamily\raggedright\setlength{\parindent}{0pt}\setlength{\parskip}{3pt}%
      \textbf{Model:}%
      \hspace{-1pt}%
      \raisebox{-0.25\height}{\includegraphics[height=1.2em]{figures/Chatgpt.png}}%
      \hspace{4pt}%
      \textbf{GPT-5.1}\par
      \textbf{Prediction (compressed):}\par
      Answer: B. \par
      - In the front view, the upcoming lane curves to the right, and the ego vehicle appears aligned with that curve, indicating a right steering input.\par
      - Several pedestrians are very close to the vehicle, and another car is close behind in the back view.\ldots
      }
      &
      {\ttfamily\raggedright\setlength{\parindent}{0pt}\setlength{\parskip}{3pt}%
      \textbf{Model:}%
      \raisebox{-0.33\height}{\includegraphics[height=2.0em]{figures/Drive-KD.png}}%
      \hspace{-3.5pt}%
      \ \textbf{Ours (Drive-KD)}\par
      \textbf{Prediction:}\par
      Final: A. \par
      Support: The ego vehicle is positioned in the center of the lane with no visible steering or trajectory deviation; the road ahead is clear and straight, and there are no indications of motion blur or speed change in the surrounding environment.\ldots\par
      }
    \end{tabular}

    \vspace{0.8em}
    \hrule
    \vspace{0.8em}

    {\ttfamily\raggedright\setlength{\parindent}{0pt}\setlength{\parskip}{3.5pt}%
    \begin{tabular}{@{}p{\linewidth}@{}}
\textbf{Evaluator:}%
\hspace{0.6em}%
\raisebox{-0.25\height}{\includegraphics[height=1.25em]{figures/deepseek-color.png}}%
\hspace{0.6em}%
\textbf{DeepSeek-V3.2}\\
\textbf{Ground Truth:} A\\[3pt]

\textbf{DeepSeek grading for GPT-5.1:}\\
1.\ Answer Correctness (50 points): 0\par
2.\ Behavioral Understanding and Detail (15 points): 0\par
3.\ Reasoning and Justification (15 points): 0\par
4.\ Contextual Relevance (10 points): 0\par
5.\ Clarity and Grammar (10 points): 10\par
Total Score: 10\par
Brief Summary: \ldots\par

\textbf{DeepSeek grading for Ours (Drive-KD):}\\
1.\ Answer Correctness (50 points): 50\par
2.\ Behavioral Understanding and Detail (15 points): 15\par
3.\ Reasoning and Justification (15 points): 15\par
4.\ Contextual Relevance (10 points): 10\par
5.\ Clarity and Grammar (10 points): 10\par
Total Score: 100\par
Brief Summary: The predicted answer exactly matches the correct answer (A), earning full points for correctness. The explanation accurately captures the ego vehicle's behavior of going straight and not moving, with clear details about lane positioning, lack of steering deviation, and environmental cues like a clear road and no motion blur. Reasoning is logical and well-supported \ldots\par
    \end{tabular}%
    }

    \medskip\hrule
  \end{minipage}
  }

  \caption{Illustration of our evaluation process on a DriveBench reasoning question. For readability, the model responses and DeepSeek justifications are compressed while preserving the grading structure and key reasons.}
  \label{fig:appendix_eval_process_example_reasoning}
\end{figure}

\begin{figure}[!t]
  \centering
  \setlength{\fboxsep}{7pt}
  \setlength{\fboxrule}{0.4pt}

  \fbox{%
  \begin{minipage}{\dimexpr\textwidth-2\fboxsep-4\fboxrule-3pt\relax}
    \footnotesize

    \textbf{Evaluation Example (Planning)}\par
    \medskip\hrule\medskip

    \centering
    \begin{minipage}{0.45\textwidth}\centering
      \includegraphics[width=\linewidth]{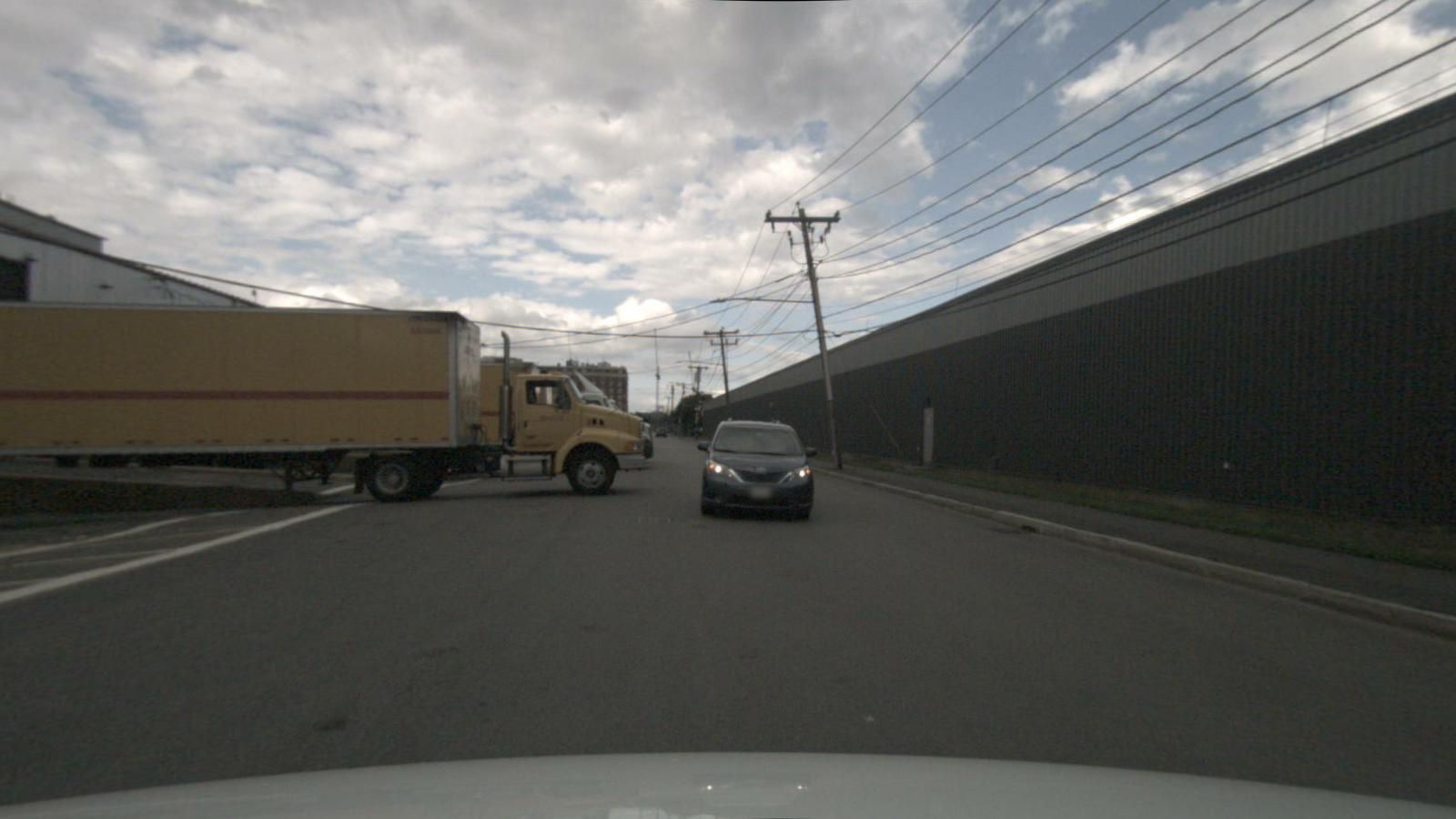}\\[-1mm]
      {\scriptsize \textbf{CAM\_BACK}}
    \end{minipage}

    \vspace{0.8em}
    \hrule
    \vspace{0.8em}

    {\ttfamily\raggedright\setlength{\parindent}{0pt}\setlength{\parskip}{4pt}%
    \begin{tabular}{@{}p{\linewidth}@{}}
\textbf{Question:} What actions could the ego vehicle take based on \texttt{<c1,CAM\_BACK,0.5073,0.5778>}? Why take this action and what's the probability?
    \end{tabular}%
    }

    \vspace{0.6em}
    \hrule
    \vspace{0.6em}

    \begin{tabular}{@{}p{0.485\textwidth}@{\hspace{0.02\textwidth}}p{0.485\textwidth}@{}}
      {\ttfamily\raggedright\setlength{\parindent}{0pt}\setlength{\parskip}{3pt}%
      \textbf{Model:}%
      \hspace{-1pt}%
      \raisebox{-0.25\height}{\includegraphics[height=1.2em]{figures/Chatgpt.png}}%
      \hspace{4pt}%
      \textbf{GPT-5.1}\par
      \textbf{Prediction (compressed):}\par
      Object: \texttt{<c1,CAM\_BACK,0.5073,0.5778>} (passenger car), behind the ego and closing.\par
      Action: maintain current lane; keep speed smooth; avoid hard braking.\par
      Reason: close follower $\rightarrow$ reduce rear-end risk; suggests cautious anticipation for potential cross-traffic / distant vehicles\ldots\par
      Probability: 0.8.\par
      Final: maintain lane with smooth, anticipatory speed control.
      }
      &
      {\ttfamily\raggedright\setlength{\parindent}{0pt}\setlength{\parskip}{3pt}%
      \textbf{Model:}%
      \raisebox{-0.33\height}{\includegraphics[height=2.0em]{figures/Drive-KD.png}}%
      \hspace{-3.5pt}%
      \ \textbf{Ours (Drive-KD)}\par
      \textbf{Prediction:}\par
      Object: \texttt{<c1,CAM\_BACK,0.5073,0.5778>} (gray sedan), relative to ego: back-left. \par
      Action: keep going at the same speed \par
      Reason: The gray sedan is in the adjacent lane moving straight ahead with no indication of braking or lane change; the road is clear ahead and no obstacles or hazards are visible in the immediate vicinity. \par
      Probability: high \par
      Final: keep going at the same speed to maintain safe clearance and traffic flow
      }
    \end{tabular}

    \vspace{0.8em}
    \hrule
    \vspace{0.8em}

    {\ttfamily\raggedright\setlength{\parindent}{0pt}\setlength{\parskip}{3.5pt}%
    \begin{tabular}{@{}p{\linewidth}@{}}
\textbf{Evaluator:}%
\hspace{0.6em}%
\raisebox{-0.25\height}{\includegraphics[height=1.25em]{figures/deepseek-color.png}}%
\hspace{0.6em}%
\textbf{DeepSeek-V3.2}\\
\textbf{Ground Truth:} keep going at the same speed; no safety issue; probability high.\\[3pt]

\textbf{DeepSeek grading for GPT-5.1:}\\
1.\ Action Prediction Accuracy (40 points): 30\par
2.\ Reasoning and Justification (20 points): 12\par
3.\ Probability or Confidence Level (15 points): 10\par
4.\ Contextual Awareness and Safety Considerations (15 points): 10\par
5.\ Conciseness and Clarity (10 points): 8\par
Total Score: 70\par
Brief Summary: \ldots\par

\textbf{DeepSeek grading for Ours (Drive-KD):}\\
1.\ Action Prediction Accuracy (40 points): 40\par
2.\ Reasoning and Justification (20 points): 20\par
3.\ Probability or Confidence Level (15 points): 15\par
4.\ Contextual Awareness and Safety Considerations (15 points): 15\par
5.\ Conciseness and Clarity (10 points): 10\par
Total Score: 100\par
Brief Summary: The predicted answer correctly identifies the action as ``keep going at the same speed,'' matching the ground truth exactly. The reasoning is clear, logical, and aligns with safety considerations, noting the absence of immediate hazards and the sedan's stable behavior in an adjacent lane. The probability is appropriately stated as ``high,'' consistent with the ground truth's ``high'' probability. \ldots\par

    \end{tabular}%
    }

    \medskip\hrule
  \end{minipage}
  }

  \caption{Illustration of our evaluation process on a DriveBench planning question. For readability, the model responses and DeepSeek justifications are compressed while preserving the grading structure and key reasons.}
  \label{fig:appendix_eval_process_example}
\end{figure}

We provide concrete evaluation-case illustrations, clarifying how our protocol is applied in practice.
Specifically, \cref{fig:appendix_eval_process_example_perception,fig:appendix_eval_process_example_reasoning,fig:appendix_eval_process_example} present representative cases for perception, reasoning, and planning, respectively.


\section{Additional Experimental Evidence}
\label{app:addition}
\subsection{AutoDriDM benchmark results}
\label{app:autodridm}

We further evaluate InternVL3-1B on AutoDriDM~\cite{tang2026drivep2dprogressiveperceptiontodecisionbenchmark}, an autonomous-driving VLM benchmark for explainable decision making.
We compare the base model, single-capability SFT models, single-capability distillation models, and the multi-teacher Drive-KD model.
Table~\ref{tab:autodridm_results} reports the corresponding results on AutoDriDM.

\begin{table}[!t]
  \caption{AutoDriDM results for InternVL3-1B variants (\%).}
  \label{tab:autodridm_results}
  \centering
  \begin{small}
    \setlength{\tabcolsep}{8pt}
    \begin{tabular}{lcccc}
      \toprule
      Model & Obj. & Scn. & Dec. & Avg. \\
      \midrule
      Base & 39.85 & 12.83 & 29.39 & 27.36 \\
      SFT-Perc. & 45.80 & 40.96 & 28.24 & 38.33 \\
      Single-Perc. & 48.73 & 46.17 & 26.96 & 40.62 \\
      SFT-Reas. & 40.30 & 20.42 & 27.94 & 29.55 \\
      Single-Reas. & 44.19 & 17.96 & 30.63 & 30.93 \\
      SFT-Plan. & 43.74 & 27.79 & 37.89 & 36.47 \\
      Single-Plan. & 46.11 & 31.25 & 40.18 & 39.18 \\
      Multi & \underline{\textbf{47.67}} & \underline{\textbf{46.81}} & \underline{\textbf{41.45}} & \underline{\textbf{45.31}} \\
      \bottomrule
    \end{tabular}
  \end{small}
\end{table}

Consistent with DriveBench, distillation generally improves over the corresponding SFT baselines under single-capability training, and multi-teacher distillation gives the best average score.
These results support the effectiveness of Drive-KD.

\subsection{Comparison with representative distillation baselines}
\label{app:representative_baselines}

We compare Drive-KD with two recent VLM/MLLM distillation baselines, EPIC/PCD~\cite{wen2025efficientmultimodallargelanguage} and Align-KD~\cite{feng_align-kd_2024}.
EPIC/PCD uses progressive consistency distillation for efficient multimodal large language models, while Align-KD transfers cross-modal alignment knowledge from a larger VLM to a smaller one.
We adapt these baselines only where required for model compatibility, train them on the same autonomous-driving data, and evaluate them with the same protocol.
Table~\ref{tab:representative_kd_baselines} reports the comparison results under this setting.

\begin{table}[!t]
  \caption{Comparison with representative distillation baselines on InternVL3-1B using DriveBench (\%).}
  \label{tab:representative_kd_baselines}
  \centering
  \begin{small}
    \setlength{\tabcolsep}{6pt}
    \begin{tabular}{lcccc}
      \toprule
      Model & Perception & Reasoning & Planning & Avg. \\
      \midrule
      InternVL3-1B & 33.26 & 20.96 & 22.36 & 25.53 \\
      Ours (InternVL3-1B, Single) & 43.13 & \underline{\textbf{34.32}} & 52.97 & 43.47 \\
      Ours (InternVL3-1B, Multi) & \underline{\textbf{43.50}} & 33.15 & \underline{\textbf{55.51}} & \underline{\textbf{44.05}} \\
      EPIC/PCD (InternVL3-1B) & 38.76 & 32.56 & 33.77 & 35.03 \\
      Align-KD (InternVL3-1B) & 37.31 & 30.19 & 31.20 & 32.90 \\
      \bottomrule
    \end{tabular}
  \end{small}
\end{table}

Drive-KD obtains the best overall performance under the same driving setting, indicating stronger transfer of driving-specific capabilities than these representative KD baselines.

\section{Additional Ablation Details}
\label{app:attn_variants}

\subsection{Per-capability layer ablations}
\label{app:layer_ablation}

We further evaluate per-capability single-teacher distillation across different layer choices on InternVL3-1B.
Each row trains on the corresponding capability split and reports the score of that capability, using the same DriveBench evaluation protocol as the main experiments.
Table~\ref{tab:per_capability_layer_ablation} reports the resulting per-capability layer ablations.

\begin{table}[!t]
  \caption{Per-capability layer ablations for InternVL3-1B single-teacher distillation on DriveBench (\%).}
  \label{tab:per_capability_layer_ablation}
  \centering
  \begin{small}
    \setlength{\tabcolsep}{6pt}
    \begin{tabular}{lccc}
      \toprule
      Setting & P & R & Pl \\
      \midrule
      CE + $A_{t\!-\!v}(1)$ & \underline{\textbf{43.13}} & -- & -- \\
      CE + $A_{t\!-\!v}(2)$ & 41.81 & -- & -- \\
      CE + $A_{t\!-\!v}(3)$ & 41.84 & -- & -- \\
      CE + $A_{t\!-\!v}(\ell_{\mathrm{pen}})$ & 40.85 & -- & -- \\
      \midrule
      CE + FA-mid $(1\to\ell_{\mathrm{pen}})$ & -- & \underline{\textbf{34.32}} & -- \\
      CE + FA-mid $(2\to\ell_{\mathrm{pen}}-1)$ & -- & 31.43 & -- \\
      CE + FA-mid $(3\to\ell_{\mathrm{pen}}-2)$ & -- & 31.22 & -- \\
      \midrule
      CE + $A_{t\!-\!v}(\ell_{\mathrm{pen}}-1)$ & -- & -- & 51.16 \\
      CE + $A_{t\!-\!v}(\ell_{\mathrm{pen}})$ & -- & -- & \underline{\textbf{52.97}} \\
      CE + $A_{t\!-\!v}(N)$ & -- & -- & 51.50 \\
      CE + $A_{t\!-\!v}(1)$ & -- & -- & 49.36 \\
      \bottomrule
    \end{tabular}
  \end{small}
\end{table}

The best perception result is obtained by early text-to-vision attention at Layer~1, while planning is strongest at the penultimate layer.
For reasoning, the broad intermediate full-attention range performs best, supporting the design that reasoning should connect early perception cues with later decision-related features.

\subsection{Additional attention distillation variants}

This appendix reports supplementary ablation results. 
We compare two types of variants against our final single-teacher recipe. 
First, we test alternative attention targets that deviate from our recipe: using full attention at Layer~1 for perception, using text-to-vision attention ($A_{t\!-\!v}$) in intermediate layers for reasoning, and using full attention at $\ell_{\mathrm{pen}}$ for planning. 
Second, we test similarity-based loss as a replacement for MSE. 
Table~\ref{tab:appendix_attn_variants} reports the corresponding extended comparisons.

We consider two attention targets: (i) full attention, which matches all query--key entries of the head-mean attention matrix, and (ii) text-to-vision attention $A_{t\!-\!v}$, which matches the submatrix whose queries are text tokens and keys are vision tokens.
In all attention-based distillation losses, we apply the standard attention masks (causal and padding) and compute the objective only on valid query--key positions shared by the teacher and student sequences.
For $A_{t\!-\!v}$, we define $T_b$ as the set of all text-query token positions in the concatenated prompt+answer sequence (excluding special and padded tokens), and define $V_b$ as the set of all vision-key token positions.
For similarity-based loss, we define cosine-based attention matching as a replacement for MSE.

\paragraph{Perception / planning (single-layer $A_{t\!-\!v}$ cosine loss).}
We apply a cosine-based loss to the text-to-vision attention $A_{t\!-\!v}$ at Layer~$1$ and $\ell_{\mathrm{pen}}$, respectively.
For sample $b$, let $T_b$ be the index set of text-query positions and $V_b$ be the index set of vision-key positions.
Define the vectorization operator $\mathrm{vec}_{t\!-\!v}(\bar A^{(\ell)}_{(\cdot),b})\in\mathbb{R}^{|T_b||V_b|}$ that stacks the entries $\{\bar A^{(\ell)}_{(\cdot),b,i,j}\}_{i\in T_b,\,j\in V_b}$.
We then apply $L_2$ normalization and compute cosine dissimilarity:
\begin{equation}
\widetilde{\mathbf a}^{(\ell)}_{(\cdot),b,t\!-\!v}
=\mathrm{vec}_{t\!-\!v}\!\left(\bar A^{(\ell)}_{(\cdot),b}\right),\qquad
\widehat{\mathbf a}^{(\ell)}_{(\cdot),b,t\!-\!v}
=\frac{\widetilde{\mathbf a}^{(\ell)}_{(\cdot),b,t\!-\!v}}
{\left\|\widetilde{\mathbf a}^{(\ell)}_{(\cdot),b,t\!-\!v}\right\|_2+\epsilon},
\end{equation}
where $\epsilon$ is a small constant for numerical stability.
The cosine loss is
\begin{equation}
\mathcal{L}_{\mathrm{att}}^{\mathrm{sim}\text{-}t\!-\!v}(\ell)
=\operatorname{mean}_{b}\left[
1-\cos\!\left(
\widehat{\mathbf a}^{(\ell)}_{s,b,t\!-\!v},
\widehat{\mathbf a}^{(\ell)}_{t,b,t\!-\!v}
\right)\right],
\end{equation}
where $\cos(\mathbf{u},\mathbf{v})=\mathbf{u}^\top\mathbf{v}$ for $L_2$-normalized vectors.

\paragraph{Reasoning (grouped intermediate layers; full-attention cosine loss).}
For intermediate-layer reasoning supervision, teacher and student may have different depths.
We therefore reuse the layer-group matching in \cref{sec:single_recipe}:
let the selected student layer set be $\mathcal{S}$ and map each $\ell\in\mathcal{S}$
to a nearby teacher-layer group $\mathcal{G}(\ell)$.
We form the teacher target as the group-mean head-mean attention:
\begin{equation}
\mu^{(\ell)}_{t,b,i,j}
=\operatorname{mean}_{k\in\mathcal{G}(\ell)} \bar A^{(k)}_{t,b,i,j}.
\end{equation}
Let $\mathrm{vec}_{\mathrm{full}}(\cdot)$ stack all entries of a head-mean full attention matrix.
We define the student/teacher attention vectors and apply $L_2$ normalization:
\begin{equation}
\widetilde{\mathbf a}^{(\ell)}_{s,b,\mathrm{full}}
=\mathrm{vec}_{\mathrm{full}}\!\left(\bar A^{(\ell)}_{s,b}\right),\qquad
\widetilde{\mathbf a}^{(\ell)}_{t,b,\mathrm{full}}
=\mathrm{vec}_{\mathrm{full}}\!\left(\mu^{(\ell)}_{t,b}\right),
\end{equation}
\begin{equation}
\widehat{\mathbf a}^{(\ell)}_{(\cdot),b,\mathrm{full}}
=\frac{\widetilde{\mathbf a}^{(\ell)}_{(\cdot),b,\mathrm{full}}}
{\left\|\widetilde{\mathbf a}^{(\ell)}_{(\cdot),b,\mathrm{full}}\right\|_2+\epsilon}.
\end{equation}
The grouped cosine distillation loss is
\begin{equation}
\mathcal{L}_{\mathrm{att}}^{\mathrm{sim}\text{-}\mathrm{full}}
=\operatorname{mean}_{b}\left[
\frac{1}{|\mathcal{S}|}\sum_{\ell\in\mathcal{S}}
\left(1-\cos\!\left(
\widehat{\mathbf a}^{(\ell)}_{s,b,\mathrm{full}},
\widehat{\mathbf a}^{(\ell)}_{t,b,\mathrm{full}}
\right)\right)
\right].
\end{equation}

\begin{table}[!t]
  \caption{InternVL3-1B extended comparisons on DriveBench (\%). 
  Each row targets one capability; other entries are not applicable (``--''). For Ours (single-teacher), each reported score is obtained by distilling and evaluating that capability using its corresponding data only, consistent with the single-capability setting above.} 
  \label{tab:appendix_attn_variants}
  \centering
  \begin{small}
    \setlength{\tabcolsep}{5pt}
    \begin{tabular}{lccc}
      \toprule
      Variant & Perception & Reasoning & Planning \\
      \midrule
      CE + Full Attn (1)                         & 42.46 & --    & --    \\
      CE + $A_{t\!-\!v}$ (mid)                   & --    & 30.42 & --    \\
      CE + Full Attn ($\ell_{\mathrm{pen}}$)     & --    & --    & 51.47 \\
      \midrule
      CE + $A_{t\!-\!v}$ (1), cosine             & 41.85 & --    & --    \\
      CE + Full Attn (mid), cosine               & --    & 32.87 & --    \\
      CE + $A_{t\!-\!v}$ ($\ell_{\mathrm{pen}}$), cosine & -- & -- & 51.76 \\
      \midrule
      Ours (single-teacher)                      & \underline{\textbf{43.13}} & \underline{\textbf{34.32}} & \underline{\textbf{52.97}} \\
      Ours (multi-teacher)                       & \underline{\textbf{43.50}} & \underline{\textbf{33.15}} & \underline{\textbf{55.51}} \\
      \bottomrule
    \end{tabular}
  \end{small}
\end{table}

As shown in Table~\ref{tab:appendix_attn_variants}, we find that alternative attention choices are consistently weaker than our final single-teacher recipe on the targeted capability. 
Finally, MSE attention matching is generally stronger than similarity-based matching in this setting, so we use MSE as the default attention distillation loss throughout Drive-KD. 

\section{Existing Assets and Terms}
\label{app:existing_assets}

\paragraph{Datasets and benchmarks.}
Our 10,500 driving distillation dataset is derived from images sampled from nuScenes~\cite{Caesar_2020_CVPR} and BDD100K~\cite{bdd100k}. The relevant upstream terms are the nuScenes Terms of Use and the BDD100K BSD 3-Clause license. DriveBench~\cite{xie2025drivebench} is our main evaluation benchmark; we use its public main branch, commit range \texttt{5eb807b--5f9ac57}, under Apache-2.0 with upstream terms applying where relevant. AutoDriDM~\cite{tang2026drivep2dprogressiveperceptiontodecisionbenchmark} is used as an additional autonomous-driving decision-making benchmark under Apache-2.0 with upstream terms applying where relevant. Vision-Flan~\cite{xu2024visionflanscalinghumanlabeledtasks}, specifically \texttt{vision-flan\_191-task\_1k}, is used as generic multimodal QA data for the pre-study analysis; its underlying source-dataset licenses apply.

\paragraph{Pretrained and API models.}
We use InternVL3 checkpoints from OpenGVLab~\cite{chen2024internvl} as students, teachers, scaling variants, and baselines. The InternVL project is MIT-licensed, and the license of the Qwen2.5 component applies where relevant. We use Qwen2.5-VL checkpoints~\cite{qwen2.5-VL} as students, teachers, and baselines; the 7B and 32B checkpoints are under Apache-2.0, while the 3B checkpoint is under the Qwen Research License Agreement and the 72B checkpoint is under the Qwen License Agreement. We use Llama-3.2-11B-Vision-Instruct and Llama-3.2-90B-Vision-Instruct as baselines under the Llama 3.2 Community License and Acceptable Use Policy. DeepSeek-V3.2~\cite{liu2025deepseek} is used through the official API as the LLM-based evaluator under the DeepSeek Open Platform Terms of Service. GPT-5.1 is used as a proprietary API baseline through OpenRouter model ID \texttt{openai/gpt-5.1}, subject to OpenRouter terms and applicable provider or model terms. We also use the publicly released ReasonDrive baseline model weights \texttt{ac4462/Qwen2.5-VL-3B-DriveLM} from Hugging Face under the Apache-2.0 license for comparison only, without redistributing the weights.


\paragraph{Baseline code.}
We compare with EPIC/PCD~\cite{wen2025efficientmultimodallargelanguage} using the official repository at commit \texttt{b2ed9cd} under Apache-2.0. We also compare with Align-KD~\cite{feng_align-kd_2024} using the official repository at commit \texttt{cfb020d}; no explicit upstream license was specified at the time of use, so we use it only for internal research comparison, do not redistribute code derived from it, and credit the original work.




\newpage

\end{document}